\documentclass[12pt, draftclsnofoot, onecolumn]{IEEEtran}

\usepackage{graphicx,color}
\usepackage{amsmath, amsthm, amsfonts, amssymb, amsbsy,nccmath}
\usepackage{mathtools}

\usepackage{algorithm}
\usepackage{enumerate}
\usepackage{lipsum}
\newtheorem{lemma}{Lemma}
\usepackage{textgreek}
\usepackage{textcomp}
\usepackage{xr}

\usepackage[sort,compress]{cite}
\usepackage{epsfig}
\usepackage{epstopdf}
\usepackage{mathtools}
\usepackage{dsfont}
\usepackage{epstopdf}

\usepackage{algpseudocode}

\usepackage{sidecap, caption}
\usepackage{subcaption}
\theoremstyle{definition}
\newtheorem{definition}{Definition}

\theoremstyle{assumption}

\theoremstyle{proposition}

\theoremstyle{corollary}

\usepackage[inline]{enumitem}   
\makeatletter
\newcommand{\inlineitem}[1][]{%
\ifnum\enit@type=\tw@
    {\descriptionlabel{#1}}
  \hspace{\labelsep}%
\else
  \ifnum\enit@type=\z@
       \refstepcounter{\@listctr}\fi
    \quad\@itemlabel\hspace{\labelsep}%
\fi}
\makeatother
\newcommand\norm[1]{\left\lVert#1\right\rVert}

\newtheorem{theorem}{Theorem}

\newcommand{\beq}{\begin{equation}}
\newcommand{\eeq}{\end{equation}}

\newcommand{\mD}{\mbox{$\mathcal D$}}
\newcommand{\mP}{\mbox{$\mathcal P$}}

\newcommand{\bmn}{\boldsymbol{n}}

\newcommand{\bmO}{\boldsymbol{O}}

\newcommand{\mS}{\mathcal{S}}
\newcommand{\mO}{\mathcal{O}}
\newcommand{\mF}{\mathcal{F}}

\newcommand{\mA}{\mathcal{A}}
\newcommand{\mE}{\mathcal{E}}
\newcommand{\mEh}{\widehat{\mE}}
\newcommand{\mR}{\mathcal{R}}
\newcommand{\mV}{\mathcal{V}}
\newcommand{\mW}{\mathcal{W}}

\newcommand{\mT}{\mathcal{T}}
\newcommand{\mM}{\mathcal{M}}

\newcommand{\mMh}{\widehat{\mathcal{M}}}

\newcommand{\mX}{{\mathcal X}}

\newcommand{\mU}{{\mathcal U}}

\newcommand{\bmd}
{\boldsymbol{d}}

\def\adots{\mathinner{\mskip0mu\raise0pt\vbox{\kern7pt\hbox{.}}\mskip3mu
          \raise4pt\hbox{.}\mskip3mu\raise8pt\hbox{.}\mskip0mu}}

\DeclareMathOperator*{\argmax}{arg\,max}
\DeclareMathOperator*{\argmin}{arg\,min}

\newcommand{\bmV}{{\boldsymbol V}}

\newcommand{\tr}{\mbox{tr}}

\newcommand{\bmh}{\bfh}

\usepackage{bm}

\newcommand{\bme}{{\bm e}}
\newcommand{\bmx}{{\bm x}}
\newcommand{\bmy}{{\bm y}}
\newcommand{\bmv}{{\bm v}}

\newcommand{\bmG}{{\bm G}}
\newcommand{\bmH}{{\bm H}}

\newcommand{\bmu}{{\bm u}}
\newcommand{\mbI}{{\mathbb I}}
\newcommand{\mbE}{{\mathbb E}}

\newcommand{\bmf}{{\bm f}}
\renewcommand{\bmh}{{\bm h}}

\newcommand{\bmz}{\bm z}

\newcommand{\bms}{{\bm s}}

\newcommand{\bmA}{{\bm A}}
\newcommand{\bmM}{{\bm M}}

\newcommand{\bmHh}{\widehat{\bmH}}

\newcommand{\bmm}{{\boldsymbol {m}}}

\newcommand{\bma}{{\bm a}}

\newcommand{\bmU}{\bm U}

\newcommand{\bTheta}{\boldsymbol{\Theta}}
\newcommand{\btheta}{\boldsymbol{\theta}}

\newcommand{\bPsi}{\boldsymbol{\Psi}}

\newcommand{\bthetah}{\widehat \btheta}

\makeatletter
\newcommand\fs@spaceruled{\def\@fs@cfont{\bfseries}\let\@fs@capt\floatc@ruled
  \def\@fs@pre{\vspace{0.5\baselineskip}\hrule height.8pt depth0pt \kern2pt}%
  \def\@fs@post{\kern1pt\hrule\relax}%
  \def\@fs@mid{\kern2pt\hrule\kern2pt}%
  \let\@fs@iftopcapt\iftrue}
\makeatother

\newcommand{\E}{\mbox{E}}

\newcommand{\bit}{\begin{itemize}}
\newcommand{\eit}{\end{itemize}}

\newcommand{\mK}{\mathcal{K}}
\newcommand{\mL}{\mathcal{L}}

\newcommand{\mG}{\mathcal{G}}
\newcommand{\mC}{\mathcal{C}}

\newcommand{\bmg}{{\boldsymbol{g}}}
\renewcommand{\bmf}{{\boldsymbol f}}
\renewcommand{\bmh}{{\boldsymbol h}}

\newcommand{\bpsi}{\boldsymbol{\psi}}

\newcommand{\bphi}{\boldsymbol{\phi}}

\renewcommand{\bmA}{{\boldsymbol A}}

\DeclarePairedDelimiter\abs{\lvert}{\rvert}%

\newcommand{\bmL}{{\boldsymbol L}}

\renewcommand{\bmU}{{\boldsymbol U}}

\newcommand{\bmsh}{\widehat{\bms}}

\usepackage{lipsum}

\restylefloat{algorithm}
\usepackage{geometry}
\def\@IEEEauthorblockNstyle{\scriptsize\@IEEEcompsocnotconfonly{\sffamily}\sublargesize}

\begin{document}

\title{Causal Semantic Communication for Digital Twins:\vspace{-3mm} A Generalizable Imitation Learning Approach\vspace{-5mm}}
\vspace{-4mm}\author{
\IEEEauthorblockN{
\normalsize Christo Kurisummoottil Thomas, \IEEEmembership{\normalsize Member, IEEE} and Walid Saad, \IEEEmembership{\normalsize Fellow, IEEE},\\\vspace{-2mm} and Yong Xiao, \IEEEmembership{\normalsize Senior Member, IEEE}
\vspace{0mm}}\\
\vspace{-3mm}
\thanks{Christo Kurisummoottil Thomas and Walid Saad are with the Wireless@VT, Bradley Department of Electrical and Computer Engineering, Virginia Tech, Arlington, VA, USA (emails:\{christokt,walids\}@vt.edu).  Yong Xiao is with the School of Electronic Information and Communications, Huazhong University of Science and Technology, Wuhan, China (email:yongxiao@hust.edu.cn).\\ \vspace{-2mm}
 }
}
\maketitle
\vspace{-17mm}
\begin{abstract}\vspace{-5mm}

A digital twin (DT) leverages a virtual representation of the physical world, along with communication (e.g., 6G), computing (e.g., edge computing), and artificial intelligence (AI) technologies to enable many connected intelligence services. In order to handle the large amounts of network data based on digital twins (DTs), wireless systems can exploit the paradigm of semantic communication (SC) for facilitating informed decision-making under strict communication constraints by utilizing AI techniques such as causal reasoning. In this paper, a novel framework called causal semantic communication (CSC) is proposed for DT-based wireless systems. The CSC system is posed as an imitation learning (IL) problem, where the transmitter, with access to optimal network control policies using a DT, teaches the receiver using SC over a bandwidth-limited wireless channel how to improve its knowledge to perform optimal control actions. The causal structure in the transmitter's data is extracted using novel approaches from the framework of deep end-to-end causal inference, thereby enabling the creation of a semantic representation that is causally invariant, which in turn helps generalize the learned knowledge of the system to new and unseen situations. The CSC decoder at the receiver is designed to extract and estimate semantic information while ensuring high semantic reliability. The receiver control policies, semantic decoder, and causal inference are formulated as a bi-level optimization problem within a variational inference framework. This problem is solved using a novel concept called network state models, inspired from world models in generative AI, that faithfully represents the environment dynamics leading to data generation. Furthermore, the proposed framework includes an analytical characterization of the performance gap that results from employing a suboptimal policy learned by the receiver that uses the transmitted semantic information to construct a model of the physical environment. The CSC system utilizes two concepts, namely the integrated information theory principle in the theory of consciousness and the abstract cell complex concept in topology, to precisely express the information content conveyed by the causal states and their relationships. Through this analysis, novel formulations of semantic information, semantic reliability, distortion, and similarity metrics are proposed, which extend beyond Shannon's concept of uncertainty. Simulation results demonstrate that the proposed CSC system outperforms conventional wireless and state-of-the-art SC systems by achieving better semantic reliability with reduced bits and enabling better control policies over time thanks to the generative AI architecture. 
\end{abstract}

\vspace{-6mm}\section{Introduction}\vspace{-3mm}

\indent \emph{Digital twins (DTs)} are replicas of the physical world \cite{TaoTII2019} created through the use of simulation software, data analytics, and sensor data. These virtual replicas rely on real-time data and advanced algorithms to model and predict the behavior and performance of physical systems, enabling better decision-making and improved efficiency across a range of industries, such as manufacturing, healthcare, transportation, and aerospace. DTs can also be used to assist wireless systems \cite{KhanCST2022} to enable self-configuration, proactive online learning, and support for \emph{connected intelligence (CI)} applications such as haptics, brain-computer interaction, flying vehicles, extended reality (XR), and the metaverse. In order to create a DT within a wireless system (e.g., at a base station or mobile edge computing server), there is need for significant  computing resources and transmission of large volumes of data. Additionally, the DT data collected at the base station will be employed to develop a model of the physical world, which can assist in making better control decisions or aiding CI applications. 
Transmitting such a staggering amount of information back from the base station to the end users can be highly inefficient from a resource utilization perspective, leading to increased latency, high power consumption, and reduced spectrum efficiency. To meet the demands of high-rate, high-reliability, and time-criticality for the aforementioned CI applications, DT-powered 6G systems could transmit only essential information relevant to the end-user. This concept forms the basis of \emph{semantic communication (SC)} systems \cite{KountourisCommMag2021,UysalIN2022,ChaccourArxiv2022}. Clearly, integrating DTs with SC offers a more effective solution for handling large amounts of data and optimizing resource utilization. Moreover, by employing SC, DTs can leverage contextual knowledge to better interpret and process data generated from diverse sources, leading to more precise predictions for event monitoring, optimal control decisions, and improved autonomous agent capabilities. However, integrating DTs with SCs poses unique \emph{challenges} particularly on the SC side. For instance, the transmitter that manages the DT must ensure that the model is trained on representative data, which can be challenging in non-stationary and dynamic wireless environments. In addition, transmitting data from a DT-enabled transmitter to facilitate real-time prediction, control, or reconstruction tasks at the end-users can be challenging due to the need for ultra-low-latency and ultra-reliable communication. 
To address the challenges of DT-based semantic communication, one promising approach is to extract the causal structure inherent in the network data. This enables the development of an accurate physical environment model at the transmit and receive nodes with fewer training samples. By leveraging generative artificial intelligence (AI) methods, this approach ultimately enables real-time prediction, reconstruction, and control.

\vspace{-7mm}\subsection{Prior Works}
\vspace{-4mm}

Despite recent AI-based SC designs \cite{XieTSP2021,SeoMehdiArxiv2022, LiuISIT2021, FarshbafanArxiv2022,FarshbafanICC2022,SunnurArxiv2022}, these prior works failed to address a few critical aspects that hinder the integration of DT and SC. Firstly, most of these designs ignore the need for a rigorous formulation of semantics, instead considering the semantic concepts as being mapped to data via a probabilistic transformation \cite{XieTSP2021}. For a DT, this probabilistic design could lead to inaccurate physical models at the end nodes, which in turn would lead to inaccurate decisions. Second, the solutions of \cite{XieTSP2021,SeoMehdiArxiv2022, LiuISIT2021, FarshbafanArxiv2022,FarshbafanICC2022,SunnurArxiv2022} suffer from limited generalizability to scenarios not encountered during training. These approaches typically use variational auto-encoders (VAEs) or transformers without structure extraction, which may reduce the ability to handle unseen wireless environments. This lack of generalizability necessitates significant retraining efforts, leading to increased communication overheads and delays, which can hinder real-time prediction, reconstruction, and control under communication constraints for DT-based wireless systems.  
Meanwhile, the works in \cite{SunTII2021,LuTII2021,OmarICC2023,DaiTII2021,LuITJ2021,HuynhWCL2022} investigated the use of DTs to enhance the performance of wireless systems. For instance, the authors in \cite{SunTII2021} proposed a distributed DT-powered federated learning framework to support edge computing in industrial IoT, while the work in \cite{LuTII2021} suggested constructing DTs at edge networks by using blockchain and federated learning. In a recent work \cite{OmarICC2023}, the authors investigated the joint synchronization of DTs and sub-metaverses within a distributed metaverse framework. Despite being interesting, the prior art \cite{SunTII2021,LuTII2021,DaiTII2021,LuITJ2021,HuynhWCL2022,OmarICC2023} has not adequately addressed modeling the environment dynamics that capture the data observed at the DT side. The only exception is the work in \cite{RuahArxiv2023} that proposed the use of a DT that builds a Bayesian model of the communication system environment using conventional signal processing techniques. 
While the Bayesian modeling approach presented is intriguing, the solution of \cite{RuahArxiv2023} is not generalizable, i.e., it is limited to the specific data distribution used to create the DT, and thus it cannot generalize to multiple wireless environment. Moreover, an important aspect missing in the prior art \cite{SunTII2021,LuTII2021,DaiTII2021,LuITJ2021,HuynhWCL2022,RuahArxiv2023,OmarICC2023} is the efficient transmission of network state and control information from DT-based nodes to edge users, which poses a challenge to achieving the high-reliability and low-latency goals of future wireless networks. To the authors' knowledge, this work will be the \emph{first} to combine SC and DTs to address the aforementioned challenges.

\vspace{-6mm}\subsection{Contributions}
\vspace{-3mm}

In contrast to the state-of-art that lacks a rigorous definition of semantics and generalizable SC system design, the main contribution of this paper is a novel framework for designing a causal SC (CSC) system consisting of two components: 1) a transmitter (called teacher) based on DTs that identifies the semantic content elements (SCEs) present in the data and performs causal discovery of the state transitions being transmitted, and 2) a receiver node (acting as apprentice) that learns the environment dynamics using the history of transmitted semantic representations and designs better control policies to maximize semantic effectiveness. While imitation learning (IL) \cite{TakayukiFTR2018} is a promising method for implementing this envisioned teacher-apprentice framework for SC \cite{XiaoJSAC2023}, practical challenges arise when implementing existing IL methods \cite{CohenArxiv2022} in DT-based SC systems. These challenges stem from the resource-intensive nature of demonstration data available at the DT side and the imperfections that may arise when transmitting data over a wireless channel, leading to inaccurate policy learning. The authors in \cite{XiaoJSAC2023} used model-free reinforcement learning (MFRL) to infer implicit semantic entities and relations from explicit semantics or observables. However, the testing performance of this approach is dependent on the specific data distribution used during training, and thus not generalizable. To overcome these issues, we propose to advance IL utilizing model-based RL (MBRL) thereby enabling the development of a model of the wireless environment dynamics using structural causal models (SCM) \cite{PearlCausality2010}. The proposed AI-based components in the SC chain rely on novel semantic information measures inspired by the concept of \emph{integrated information theory (IIT)} in the literature on the \emph{theory of consciousness} \cite{AlbantakisArxiv2022}. IIT measures help identify distinct \emph{SCEs}, which hold meanings or semantics and are present within the network states observed by the DT. In summary, our key contributions include:
\begin{itemize}
\vspace{-1mm}\item We introduce \emph{novel information measures} for the learned \emph{SCM} at the imitator, inspired from the IIT measures \cite{TononiEntropy2019}. 
We also propose a new semantic state abstraction concept that utilizes the intrinsic information concept from IIT. Semantic state abstraction plays a crucial role as it helps eliminate irrelevant information observed at the transmitter side, thus reducing transmitted information. Moreover, it enhances the ability of the agent to generalize to previously unseen areas of the state space, which could correspond to different wireless environments. This allows us to develop generalizable native AI-based wireless SC systems.\vspace{-1mm}
\item We use the concept of abstract simplical complexes in topology to theoretically characterize causal relations among SCEs in the data observed at the DT. Moreover, we demonstrate that the SCEs and their associations can be algebraically defined as a cell complex structure. By defining the SCEs in accordance with IIT, these topological characterizations enable us to define semantic metrics such as similarity and information that go beyond conventional information theory concepts. This capability enables the creation of DT-based SC systems that can precisely identify the causal structure and consequently develop causally invariant semantic representations.  \vspace{-1mm}
\item At the receiver, we create a \emph{``network state model"} similar to the concept of world models in generative AI \cite{MatsuoLeCunnNN2022}. We design a semantic decoder that extracts the maximum semantic information. However, due to limited computing resources at the imitator node, the model learns a sub-optimal version of the environment dynamics. This model is then used to learn imitator policies for controlling communication tasks. To solve for the neural network (NN) parameters, we propose a bi-level optimization method within an MBRL framework. This approach enables us to learn both the state transition model and the imitator policy. \vspace{-1mm}
\item We analytically characterize the performance shortfall (compared to the expert agent's policy) in terms of the quality-of-experience (QoE) associated with the sub-optimal network state models created at the receiver node. The QoE is defined as a function of the semantic effectiveness resulting from the transmitted semantic representation. \vspace{-1mm}
\item Simulation results demonstrate the superiority of the proposed CSC in semantic reliability and throughput delivered by a DT-based SC system, in contrast to similar wireless systems that rely on traditional AI approaches. Furthermore, even when the channel quality is very poor, the proposed CSC achieves a semantic reliability that is $3.4$ times better than classical DT-based SC systems due to its generative AI architecture at the receiver that allows it to reconstruct data. The simulations conducted in our study provide further evidence that the proposed CSC can adapt quickly to non-stationary conditions in wireless environments. The CSC requires approximately 3K fewer samples for retraining, which validates its generalizability. \vspace{-1mm}
\end{itemize}

The rest of this paper is organized as follows. In Section II, we present the proposed CSC system model. Section III describes the proposed IIT measures for quantifying the semantic information. Section IV introduces the MBRL based IL for learning the transmitter and receiver modules. Section IV provides simulation results. Finally, conclusions are drawn in Section VI\footnote{\vspace{-4mm}Appendix is provided as supplementary material.}. 

\textbf{Notations:} Lower-case letter $a$ is a scalar, boldface lower-case $\bma$ and upper-case letter $\bmA$ represent a vector and a matrix, respectively. A set (either discrete or continuous entries) or a topological space is represented using Calligraphic font $\mX$. $\mR^M$ represents an $M-$dimensional vector whose entries belong to real numbers $\mR$. $[a,b]$ represents the real number range between $a$ and $b$. $\bmA\odot \bmA$ represents the element-wise multiplication of two matrices. $\abs{a}$ represents the absolute value of $a$.

\vspace{-4mm}\section{System Model}
\vspace{-2mm}

\vspace{-2mm}\begin{figure}[t]
\centering{\includegraphics[width=7.2in,height=4.0in]{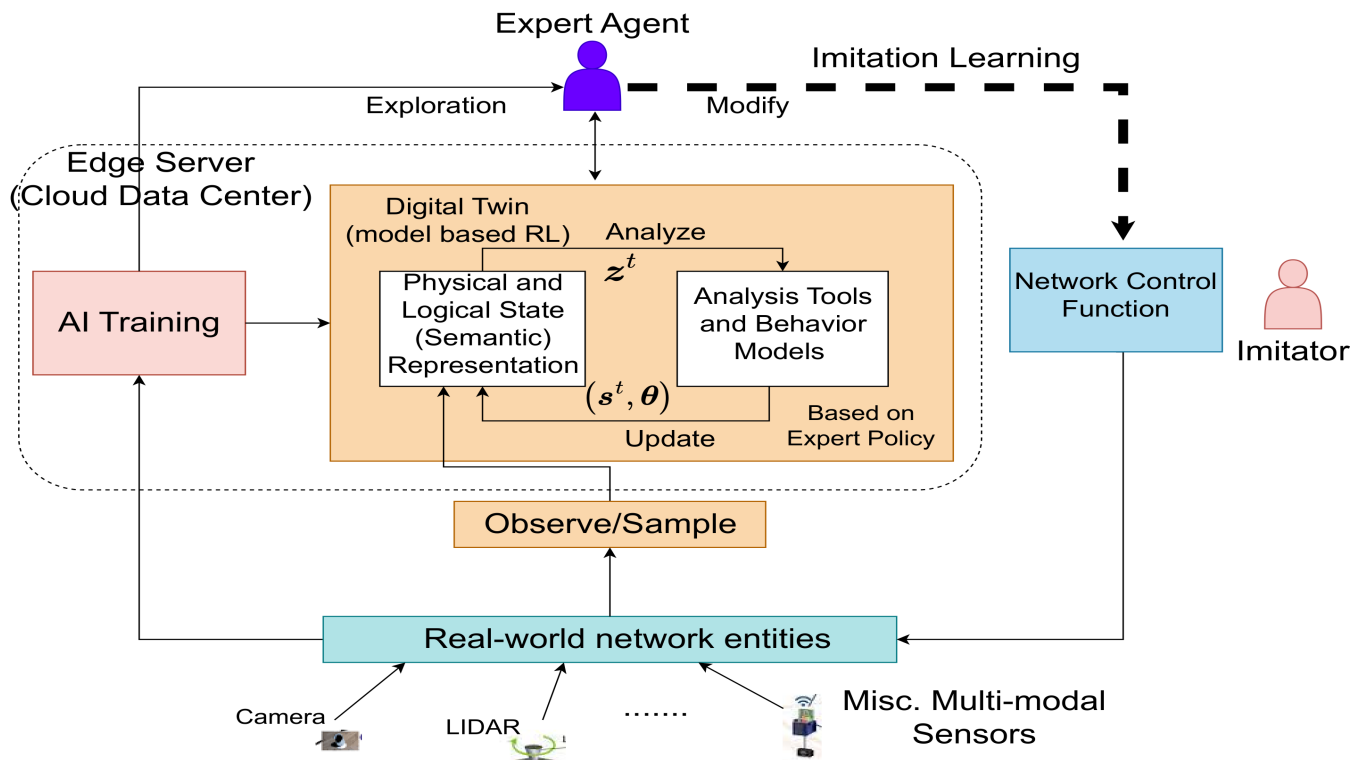}}\vspace{-3mm}
\caption{Illustration of our digital twin model showing expert (transmitter that has access to different sensor observations) and imitator agent interactions.}
\label{DTModel}\vspace{-2mm}
\vspace{-9mm}
\end{figure}
Consider a point-to-point SC system that employs a DT to aid in making control decisions. This system can capture many practical applications. For example, in the context of 6G and open radio access network (ORAN) systems, a network DT located at a cloud data center would replicate various latent aspects of a network, including signals, coverage, interference, traffic behavior, and user mobility, across different frequency layers. The DT optimizes sensitive parameters, such as radiated power or multi-user scheduling, by providing a safe simulation environment without any actual risks to the real network.
Within the ORAN architecture, the radio unit (RU) component is located at the base station and has limited computational power \cite{NguyenArxiv2023}. To compensate for this, the RU relies on the transmission of information from the control unit (CU) component that implements the DT. The CU has more knowledge about the network architecture and user profiles, aided by the cloud data center, and can transmit this information to the RU via a fronthaul wireless link. By implementing a network DT-based SC system over the fronthaul link, the network's efficiency (bandwidth utilization) and security can be significantly enhanced, making it an essential tool for network management.
Another key example that is captured by our model is in CI applications, where DTs can improve user experience in various autonomous wireless services, such as robotic surgery, telehealth, driverless vehicles, and industrial robotic manufacturing sites. By creating a virtual representation of the physical system, a DT can simulate and predict its behavior in real-time, enabling proactive decision-making and optimized system performance to enhance the user experience. For instance, a DT can reduce latency and ensure reliable communication in telehealth services, or enhance safety and efficiency in driverless vehicles and industrial robotic manufacturing sites.

\begin{figure}[t]
\centering{\includegraphics[width=6.8in,height=4.8in]{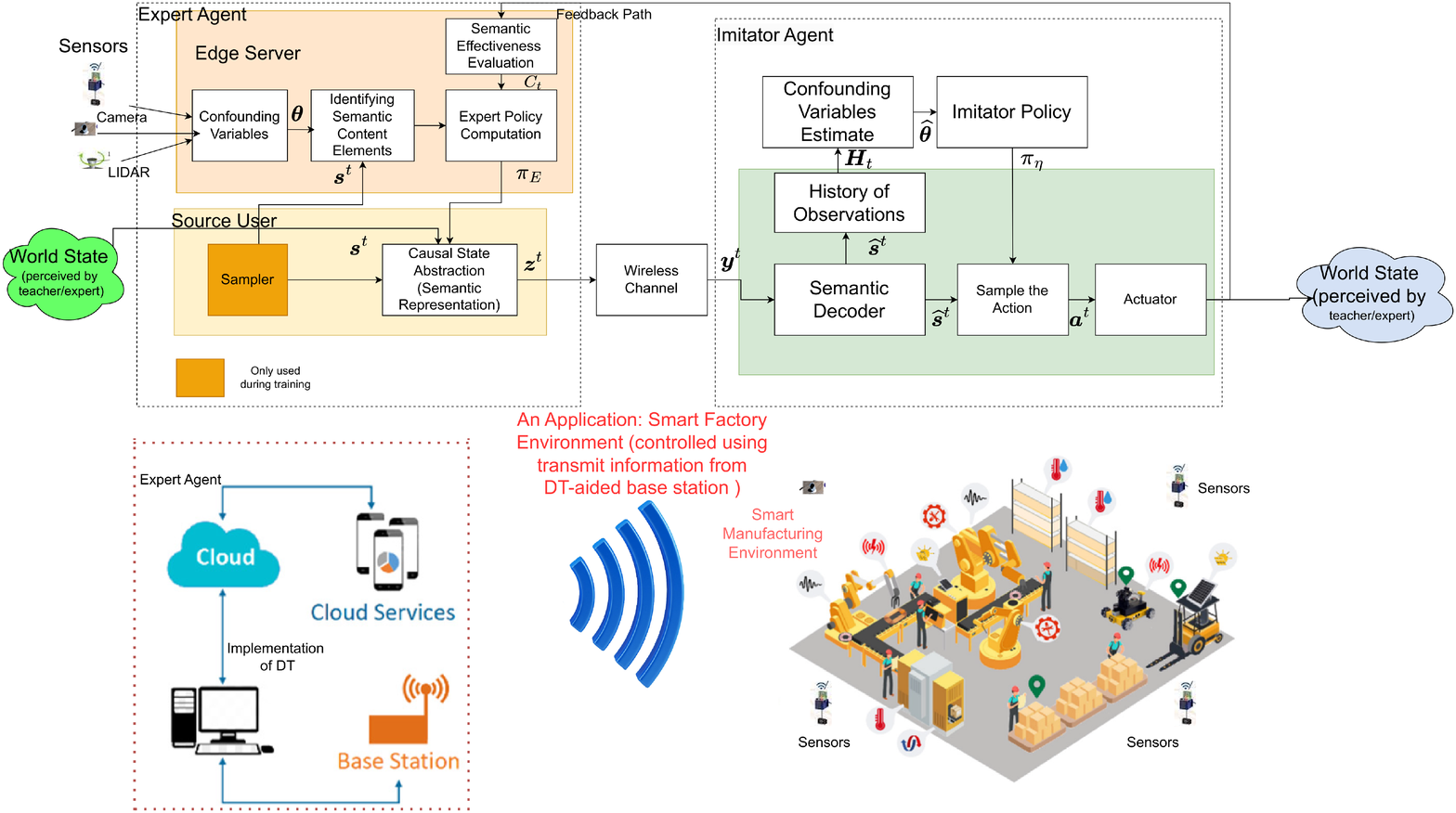}}\vspace{-15mm}
\caption{\small Proposed SC system framework based on IL. The figure also illustrates an example application as an autonomous wireless system that includes a DT designed explicitly for a smart manufacturing environment. The application is based on sensing information gathered at the central node and is tailored to the proposed system model.}
\label{SystemModel}\vspace{-1mm}
\vspace{-7mm}
\end{figure}

The functionalities of the DT are summarized in Fig.~\ref{DTModel}. The DT is created by an expert agent who analyzes observations from various sensing elements (defined as $(\bms^t,\btheta)$) in the wireless environment. These observations are used to model the physical and logical state of the network represented by the transition probability $p(\bms^{t+1}\mid\bms^t,\btheta,\bma^t)$, which is then used to formulate optimal control decisions $\bma^t$. The expert agent communicates either a condensed version of the network state or the optimal control decisions defined as $\bmz^t$ to the imitator agent. Subsequently, the imitator extracts the state information and performs the necessary low-level actions. The interactions between expert and imitator agents in the SC system, as described here, can be classified as an example of IL. 
By utilizing a DT, accurate physical system models and real-time data can be made available to inform decision-making, facilitating effective control decisions. We next explain the details of our DT-based SC system as shown in Fig.~\ref{SystemModel}. The source side of the transmission consists of a resource-constrained node aided by an edge server with significant computational and memory capabilities. The source user and the edge server together act as the expert agent (that implements the DT) to teach an apprentice or imitator agent, serving as the receiver, with no prior knowledge of the expert's data. 
Here, one naturally wonders whether communicating only the expert actions determined using the policy evaluated by the DT could be sufficient. While communicating only actions may be relevant when the imitator agent has limited capabilities, for cases in which the agent must possess more advanced intelligence, additional information beyond actions may be necessary for effective imitation. In the case of a SC system, building an intelligent imitator agent is crucial. For instance, during channel outages lasting over several communication time slots, the receiver must still generate policy information by leveraging the historical states received over the network and potentially using limited sensing information. Merely communicating the actions does not offer significant information about the observed states at the transmitter, which, in turn, fails to facilitate the environment modeling at the imitator. This, in turn, motivates designing systems that communicate meaningful semantic representations of the states observed at the expert agent to the imitator. Further, the imitator can utilize the potential of generative AI algorithms \cite{DavidNIPS2018} to reconstruct the network states. Next, we outline the different AI-based components present in the proposed SC system.

\vspace{-5mm}\subsection{Causal imitation learning for digital twin-based SC systems}
\vspace{-2mm}

 Given the IL model, we implement the SC functionalities as follows. We divide the transmissions into training and communication phases. During training, the source user samples random initial states from a particular distribution, $p(\bms^0)$. The sampler implements this functionality during training while the states are observable from the environment during the communication phase. Given the initial state, the state transition is determined by the action of the source, and it is sampled using the \emph{expert policy}, $\bma^t\sim \pi_E(\bma^t\mid\bms^{t},\btheta)$. The optimal action $\bma^t$ depends on the state at the current time $t$ and \emph{confounding variables} (from various sensing devices) $\btheta$ that could represent background information about the communication environment accessible only to the edge server. The expert agent computes the future state $\bms^{t+1}\!\sim \!p(\bms^{t+1}\mid\bms^t,\bma^t,\btheta)$, using the information from $\btheta$ and the current state $\bms_{t}$. In our DT-based system, efficient transmission is of utmost importance due to the vast amounts of data the expert agent processes. To address this issue, the expert agent could exploit a \emph{causal discovery AI} method \cite{GeffnerArxiv2022} to extract data structure (inherent in the state transition dynamics) while identifying the SCEs. In this setup, a potential question is how to identify the \emph{SCEs} (defined as $\bmu^t$) comprised in any subset of the state $\bms_i^t$. SCEs are entities in the source data that hold \emph{meaning (semantics)} or significance, aiming to improve the knowledge learned by the imitator. We aim to identify the SCEs in the data and characterize the semantic information in the causal structure. Causality is apt here since it helps with generalization in SC. It allows the communicating nodes to understand the underlying mechanisms and processes that drive the relationships between network variables rather than just observing their associations. This deeper understanding of the underlying causal relationships can help the nodes to make more accurate predictions or control decisions and develop more robust models that can generalize well to new, unseen data. To represent the causal structure that underlies the transition from $\bms^t$ to $\bms^{t+1}$, our CSC system employs an MBRL approach. The causal model underlying the state transitions based on Pearl's representation of causality \cite{PearlCausality2010} is defined next.
\vspace{-4mm}\begin{definition} 
An \emph{SCM} is a collection of elements represented as $<\bmU, \bmV, \mF, P(\bmU)>$, where $\bmV$ represents endogenous variables (cause and effect variables), and $\bmU$ represents exogenous variables (random, unknown noise). The set of structural functions $\mF$ maps each $f_v \in \mF$ to $\bmv\in \bmV$ such that $f_v(Pa_v,\bmu_v)$ determines $\bmv$ where the set of parents $Pa_v\subset \bmV$ and $\bmu_v\subset \bmU$. The exogenous distribution $P(\bmU)$ determines the values of $\bmU$, and thus the distribution of endogenous variables $\bmV.$ 
\vspace{-10mm}\end{definition}

The apprentice (imitator) can observe only a subset of endogenous variables, partitioned into $\bmO$ and $\bmL$, where $\bmO = {\bms^t}$ (in reality, an encoded version communicated by the teacher) and $\bmL = \btheta$ (learned by the apprentice). Here, $\bms_i^t \in \mR^D$. The marginal distribution $P(\bmO)$ is called the observational distribution. 
MBRL allows communicating the \emph{causal state abstraction} (i.e., the ``semantic representation") $\bmz^t$ at any given time over a wireless channel susceptible to errors. 
 The confounding variables are not transmitted due to the communication and memory constraints associated with the wireless channel and imitator, respectively.  
 The design of the semantic encoder (that maps the states to actual transmit signals) must be done in a way to reduce the amount of data transmitted while maintaining \emph{semantic effectiveness}, defined as a measure of the accuracy of the actions performed using the semantic reconstruction process at the imitator's end. 
 \subsubsection{MBRL Model} Next, we formulate the problem of expert policy computation as sequential decision-making in a finite-horizon episodic Markov decision process (MDP), defined as $\mathcal{M} = (\mS,\mA,\mR,\mT,p(\bms^0),H, \btheta)$. Here, $\btheta$ is a vector of confounding variables that the imitator does not have access to and $P(\btheta)$ is the probability distribution of $\btheta$. $\btheta$ is fixed within each episode but can vary across different episodes. The $M-$dimensional vector $\btheta$ represents the sensing
information available from $M$ sensors which may provide different modality information to
supplement the optimal policy. The MDP framework comprises a set of states $\mS$, a set of actions $\mA$, a deterministic reward function $\mR:\mS\times \mA \rightarrow [0,1]$ that provides evaluative feedback to the imitator agent, a transition function $\mT:\mS\times \mA \rightarrow \Delta(\mS)$ that describes the distribution of future states, an initial state distribution $p(\bms^0) \in \Delta(\mS)$, and a maximum episode length or horizon $H$. The unit interval constraint for the reward function is adopted for ease of analysis, and our framework can handle general reward functions. The imitator agent cannot access neither the transition function $\mT$ nor the reward function $\mR$; hence, both are treated as random variables (known as the \emph{environment}). The $N$-dimensional causal state description $\bms \in \mS \subset \mathcal{R}^N$ is extracted from raw observations and represented as an SCM. Let $\mM^{*}$ be the true MDP that the agent interacts with and attempts to solve over $K$ episodes, with the environment model $\mE = (\mR^{*}, \mT^{*},\btheta)$. However, the imitator agent's computing capabilities and time constraints limit the accuracy of the true environment that can be learned, resulting in a learned MDP $\mMh$ (the respective learned environment is $\mEh$). Each episode involves the agent taking exactly $H$ steps, starting from an initial state $\bms^0 \sim p(\bms^0)$. For each $h \in \{1,\cdots,H\}$, the agent observes the current state $\bms^h \in \mS$, selects an action $\bma^{h} \sim \pi^{(h)}(\cdot \mid \bms^h) \in \mA$, receives a reward $r_h = \mR(\bms^h,\bma^{h}) \in [0,1]$, and transitions to the next state $\bms^{h+1} \sim \mT(\cdot \mid \bms^h,\bma^{h}) \in \mS$. We let $\tau_k = (\bms^1_{(k)},\bma^1_{(k)},r^1_{(k)},\cdots,\bms^H_{(k)},\bma^H_{(k)},r^H_{(k)},\bms^{H+1}_{(k)})$ be the random variable that represents the trajectory experienced by the agent in the episode $k$. $\bmH_k = (\tau_1,\cdots,\tau_{k-1}) \in \mathcal{H}_k$ is the random variable representing the entire history of the agent's interaction with the environment at the start of the episode $k$. The regret of the RL algorithm over $K$ episodes is:
$
r_g(K,\pi^{(1)},\cdots,\pi^{(K)},\mE^{*})= \bar{V}^{*}_{\mE^*,1} - \bar{V}^{\pi^{(k)}}_{\mE^*,1},$
with the value function $\bar{V}^{\pi^{(k)}}_{\mE,h} = \mathbb{E}_{a\sim \pi^{(k)}(\cdot\mid \bms)}\left[
Q^{\pi}_{\mE,h}(s, a)\right]$. The associated action-value function is $Q^{\pi}_{\mE,h}(s, a) = \mathbb{E}\left[\sum\limits_{h^{\prime}}^{H}\mR(\bms^{h^{\prime}},\bma^{h^{\prime}})\mid \bms^h=\bms,\bma^{h}=\bma\right]$, where
the expectation integrates over randomness in the action selections and transition dynamics.
The Bayesian regret is defined as the expected value of the sum of the episodic regrets: $\bar{r}_g(K,\pi^{(1)},\cdots,\pi^{(K)},\mE^{*}) = \mathbb{E}\left[\sum_{k=1}^K \left[\bar{V}^{*}_{\mE^*,1} - \bar{V}^{\pi^{(k)}}_{\mE^*,1}\right]\right].$
Note that the regret is a random variable due to the uncertainty in $\mE$. However, we emphasize again here that using a classical MFRL algorithm is not generalizable to multiple wireless environments, which necessitates us revisiting the SC design approach using causal machine learning (ML) tools as explained in later sections. To incorporate causality, the IL framework can be reformulated using an MBRL approach, as proposed in \cite{MoerlandFTML2023}. Compared to MFRL, MBRL is generally more efficient, making it particularly well-suited for SC. This is because MBRL employs a learned model of the environment $p(\bmsh^{t+1}\mid\bmsh^t,\bma^t,\bthetah)$ (part of the imitator's job here), aided by a history of received information, to plan and update its policy. 
However, similar to other supervised learning methods, MBRL faces the challenge of generalization. The data used for training may not match the one encountered during testing, and even small inaccuracies in the dynamics model or changes in the control policy can lead to exploring new parts of the state space. Thus, we need new models that have strong generalization capabilities in MBRL. To achieve the SC goal of improved resource efficiency, we propose to bridge this gap in the literature for IL by developing generalizable (across several wireless environments) solutions that are causality aware. 

The mismatch in knowledge between expert agent ($\mE$) and imitator agent ($\mEh$) that gets learned using MBRL can lead to incorrect inferences due to auto-suggestive delusions \cite{OrtegaArxiv2021}, i.e., false beliefs generated within one's mind. To address this issue, we present a novel approach by applying a new IL principle, in which we treat actions as causal interventions, as described in \cite{PearlCausality2010}. In supervised learning, one can condition or intervene on data using factual and counterfactual error signals, respectively, which helps resolve auto-suggestive delusions. To account for imperfections in the imitator's model, we employ confounding variables in causality. We begin by using SCMs to exploit sparsity in state transitions, which helps generalize to unseen parts of the state-action space. We then extend the IL framework to include some variables $\btheta \in \bTheta$ observed by the expert but not the imitator. This setup, introduced in \cite{CohenArxiv2022} for RL, allows modeling of the unknown part of the environment that the imitator learns from its previous state-action sequences. The method in \cite{CohenArxiv2022} assumes that the imitator has perfect knowledge of $\bms^t$, which differs from our SC system. Also, their AI approach relies on conventional data-driven techniques, such as VAE (without any structure extraction), that may not be sufficient to achieve generalizable wireless systems. Given the MBRL model, in the next step, we discuss how the expert agent computes its expert policy using the state and confounding variables to which it has access. Additionally, we will explore how a causally invariant semantic language is defined.
\vspace{-1mm}\subsubsection{Expert policy and causally invariant semantic language}
The source user does not know the confounding variables because it is a computationally constrained node. In this case, the source user communicates the state information to the edge server. Along with confounding variables, the edge server computes an optimal policy, called the \emph{expert policy}, $\pi_E$. $\pi_E$ is computed to maximize the average reward over time, $\pi_E^{*} = \argmax\limits_{\pi_E} \mathbb{E}\left[\mR(\bms)\right]$, where $\mR(\bms) = \mathbb{E}_{\bma\sim \pi_E(\bma^t\mid\bms^t,\btheta)}\mR(\bms,\bma)$. $\pi_E$ is communicated to the source user via a highly reliable backhaul connection. Using $\pi_E$, the source user samples transitioned state information. Now, the source user must extract the causality between the state transitions which can be computed as the distribution (see \cite{ScholkopfPIEEE2021} for more details on causal representation as posterior factorization):
\vspace{-1mm}\beq
\begin{aligned}
\mP_1: \mbox{Compute}\,\,\,p(\bms^{t+1}\mid\bms^t,\bma^t,\btheta) = \prod\limits_i\underbrace{p(\bms_i^{t+1}\mid Pa(\bms_i^t))}_{\mbox{\small Causal graph structure}}\,\,\,\mbox{subject to} \,\,\, \bma^t = \pi_E^{*}(\bma^t\mid\bms^t,\btheta).
\end{aligned}\label{eq_P1}
\vspace{-1mm}\eeq
The factorization of the transition probability in \eqref{eq_P1} results from causal discovery. 
This concept of identifying causal features is closely linked to sparsity, which is a form of \emph{inductive bias} that shapes the behavior of the learning agent. An important aspect of this approach is the continuous improvement of the learning policy, which helps us achieve both causal invariance and sparsity \cite{TomarArxiv2021}. With this in mind, we can ask whether we can leverage the sparsity of transition dynamics to develop a model that generalizes better to unseen parts of the state-action space. Moreover, using the associational, interventional, and counterfactual levels of reasoning, the source user can generalize to distinct data distributions. Training AI models to generalize over the wireless environment, rather than specializing in fitting specific conditions (e.g., a single cell), provides the necessary robustness to deploy a single AI model for a given task across the entire network. This improves scalability and reduces the complexity (in terms of training + distinct AI models) of the AI functionalities and operations. Additionally, we establish the semantic representation that emerges from the proposed semantic language, which is defined below.
\vspace{-4mm}\begin{definition}
A \emph{semantic language} $\mL = (\bms^t,\bmz^t)$, is a mapping
from the observed states $\bms^t$, to their corresponding semantic
representation $\bmz^t$, based on the identified SCEs and the causal graph. This mapping is described as the encoder probability distribution $p(\bmz^t\mid\bms^t,\bms^{t+1})$.
\vspace{-5mm}\end{definition}
To achieve the goal of reducing transmission in the SC system, it is optimal to compute the encoder distribution by determining the NN parameters $\bPsi$ that result in the maximum received semantic information at the imitator node, as follows:
\vspace{-2mm}\beq
\begin{aligned}
\mP_2: \argmax\limits_{p_{\bPsi}\left(\bmz^t\mid (\bms^t,\bms^{t+1})\right)}  \mbI(\bms^t; \bmy^t) \\
\mbox{subject to}\,\, \mathbb{I} (\bms^t; \bmz^t)
\leq \mathbb{I}_b,
\end{aligned}
\label{eq_P2_init}
\vspace{-2mm}\eeq
 where $\mathbb{I}_b$ represents the bandwidth limitations of the wireless channel. The information measures above correspond to the novel semantic information measures detailed in Section~\ref{IIT} (see equation~\eqref{eq_Iphi_c_e} and~\eqref{eq_Iphi}). The abstracted state information $\bmz^t$ depends on two factors: 1) the IIT, which helps identify SCEs and rigorously formulate semantic information, and 2) semantic awareness, which is determined by user-defined key performance indicators (KPIs) such as QoE. We next define the QoE metric below. 

\vspace{-1mm} \subsubsection{Semantic effectiveness as QoE metric}

On the imitator's side, the semantic decoder module should be designed to maximize semantic effectiveness defined based on our work \cite{ChristoTWCArxiv2022}. We thus introduce the metric $C_t$, which captures the causal impact of the expert agent's message (via the imitator's actions) as observed through a channel with a response characterized using $p(\bmy^t\mid \bmz^t)$ (this distribution could capture the fading and interference in the wireless environment). In other words, $C_t$ measures the semantic effectiveness (inversely proportional to $C_t$) of the transmitted message to the end-user. We define 
\vspace{-2mm}
\beq
\vspace{-0mm}
\begin{array}{l} 
C_t(\bms^t,\bmsh^t)=  \textrm{KLD} \left(\pi_{E}(\bma^t \mid \bms^t,\btheta) \,\mid\mid\, \sum\limits_{\bmy^t,\bmz^t} p(\bmz^t\mid\bms^t)p({\bmy}^t \mid {\bmz}^t )p(\widehat{\bms}^t \mid {\bmy}^t )p(\bthetah^t\mid \bmH_t)\pi_{\eta}(\bma^t\mid\widehat{\bms}^t,\bthetah^t)\right) ,
\end{array}
\label{eq_sem_eff}
\vspace{-2mm}
\eeq
where $\textrm{KLD} (p\mid\mid q)$ represents the KL divergence between $p$ and $q$. $\pi_{\eta}(\bma^t\mid\widehat{\bms}^t,\bthetah^t)$ is the sub-optimal policy computed by the imitator using the limited information it received over the air. $p(\widehat{\bms}^t \mid {\bmy}^t )$ denotes the semantic decoder distribution. The decoder component can thus be designed using the game-theoretic framework we developed in \cite{ChristoTWCArxiv2022}. The second term in the quantity $C_t$ is an average measure across the channel realization. Since the semantic effectiveness metric in \eqref{eq_sem_eff} captures the channel effect on the transmitted semantics, the learned policy (for the imitator) parameters is robust to the channel errors (in terms of the best the imitator can do).
The semantic effectiveness measure is the distance in the semantic space (distance between policy distributions) and not a conventional metric like Euclidean distance. However, computing the semantic effectiveness at the imitator requires it to know $\pi_E$. This necessitates communicating the imitator policy to the edge server via the backhaul. Subsequently, the edge server assesses $C_t$ and transmits it to the imitator, which then adapts its policy based on the quality of semantic effectiveness. The communication link between the imitator and the edge server is assumed to be highly reliable compared to the wireless connection to the source user. 

Having defined semantic effectiveness, we look at how to define the reward signal mentioned earlier. The agent may encounter extremely sparse or entirely absent extrinsic rewards in numerous real-world situations \cite{PathakICML2017}. In these cases, the intrinsic reward signal based on semantic effectiveness can be used to facilitate collaborative exploration and communication skill learning (specifically, semantic encoder and decoder) between the expert and imitator agents. This skill becomes helpful in subsequent communication phases. Motivated by these considerations, we can define $r_t = r_t^i + r_t^e,$ where $r_t^e$ is the extrinsic reward which is mostly
(if not always) zero and the intrinsic reward, $r_t^i = \frac{1}{1+C_t(\bms^t,\bmsh^t)}$. By incorporating semantic effectiveness when computing $\pi_E$, the prediction of state transitions and causal structure is improved. This leads to a better design of the semantic encoder on the expert side, which adjusts its transmission strategy to maximize the semantic information extracted by the receiver. Having outlined the different IL components for CSC, we next develop a novel semantic information measure that helps us identify the SCEs and that is critical in formulating various objective functions ($\mP_1,\mP_2$ in \eqref{eq_P1} and \eqref{eq_P2_init}) elaborated in Section~\ref{CDL_IL}.

\vspace{-4mm}\section{Characterizing Semantic Concepts via Integrated Information Theory}
\label{IIT}\vspace{-2mm}

An SCM with respect to the formulations in IIT can be described as a stochastic system $\mU = \{U_1,   \cdots , U_n\}$ of $n$
interacting units (atomic) with state space $\Omega_{\mU} \!=\!
\prod_i \Omega_{U_i}$, where $\Omega_{U_i} \!\subset\! \mR$,
and the current state is $\bmu\! \in \!\Omega_{\mU}$. Each network state $\bms^t_i$ in \eqref{eq_P1} that is observed at DT can be composed of multiple such atomic units, whose state space we denote as $\Omega_{s_i}\subset \Omega_{\mU}$. Our system is updated in discrete steps such that the state space $\Omega_{\mU}$ is finite, and the
individual random variables $U_i \in {\mU}$ are conditionally independent given
the preceding state of ${\mU}$:
$p(\bmu^{t+1}\mid \bmu^t) = \prod_i p(u^{t+1}_i\mid \bmu^t).$
The IIT-based concept of semantic information draws inspiration from the theory of consciousness measures proposed by \cite{OizumiPLOS2014}. This concept is constructed on the fundamental principles of intrinsic information, information integration, and exclusion, which we elaborate on next.

\vspace{-6mm}\subsection{Intrinsic Information for State Abstraction}
\vspace{-2mm}

 In IIT, \emph{intrinsic information} refers to the inherent cause-and-effect structure within a system that produces the particular set of observed states and transitions. Put simply, it is the semantic information built into our DT-based system that leads to the shift from one state ($\bms^t$) to the next ($\bms^{t+1}$). Moreover, in IIT, information present in $\bms^t$ is considered to be causal only if it has selective causes and selective effects within the system. 
 This selective nature of information distinguishes it from mere correlation or statistical association between variables (which is the extrinsic notion of information used in Shannon's theory of communication), which may not be causal in nature. We now proceed to analytically define intrinsic information. The amount of information that the current state $\bms_i^t$ (which represents any subset of causal variables part of the SCM) specifies about the past, i.e., its cause information $\mbI_c$, is measured as the distance between the cause repertoire $p(\bms_i^{t-1}\mid \bms_i^t )$ and the unconstrained past repertoire $p(\bms_i^{t-1})$, and is defined as follows:
\vspace{-2mm}\beq\vspace{-1mm}
\begin{array}{l}
\mbI_c(\bms_i^{t-1}\mid \bms_i^t) = \mathbb{D}\left(p\left(\bms_i^{t-1}\mid \bms_i^t\right ) || p\left(\bms_i^{t-1}\right)\right).
\end{array}\vspace{-2mm}
\eeq
Just like cause information $\mbI_c$, the effect information $\mbI_e$ of $\bms_i^{t}$ is
quantified as the distance between the effect repertoire of $ \bms_i^t$ and
the unconstrained future repertoire $p(\bms_i^{t+1} )$ and written as
\vspace{-2mm}\beq
\begin{array}{l}
\mbI_e(\bms_i^{t+1}\mid\bms_i^t) = \mathbb{D}\left(p\left(\bms_i^{t+1}\mid \bms_i^t\right) || p\left(\bms_i^{t+1}\right)\right).
\end{array}\vspace{-2mm}
\eeq
Next, we define an appropriate measure for $\mathbb{D}$. The Kullback-Leibler divergence (KLD) is a useful metric from classical information theory, but it is not a true metric (it is not symmetric) and it is unbounded. Additionally, KLD only measures how ``sharp" a distribution is compared to another, without taking into account whether some states of the system are closer than others (in the sense of the Euclidean distance metric). Compared to the literature in IIT \cite{AlbantakisArxiv2022}, a more appropriate measure that aligns better with the IIT notion of information as ``differences that make a difference" is the \emph{earth mover's distance (EMD) or Wasserstein distance}. This distance is formulated in the context of optimal transport problems as follows:
\vspace{-4mm}\begin{definition}
  For any two probability measures $\mu_s,\mu_t$, the optimal cost of transporting from $\mu_s$ to $\mu_t$ can be formulated as the \emph{Wasserstein distance} $W_p^p(\mu_s,\mu_t) = \min\limits_{\gamma \in \mathcal{P}}\int\limits_{\Omega_x \times \Omega_y} \norm{x-y}^p \gamma(x,y)dx dy,$
where $\Omega_x$ and $\Omega_y$ are the domain of $\mu_s$ and $\mu_t$, respectively. $\mP$ is the set of joint probability distributions.
\vspace{-4mm}\end{definition}
$W_p^p(\mu_s,\mu_t)$ is symmetric and bounded.
Finally, having calculated $\mbI_c$ and $\mbI_e$,
the total amount of cause-effect information $\mbI_{ce}$ specified by $\bms_i^{t}$ over
the purview defined as $\{\bms_i^{t+1},\bms_i^{t-1}\}$, is: 
\vspace{-2mm}\beq
\vspace{-1mm}\begin{array}{l}
\mbI_{ce}(\bms_i^{t+1},\bms_i^{t-1}\mid \bms_i^{t}) = \min (\mbI_{c},\mbI_{e}).
\end{array}
\vspace{-1mm}\eeq
From the intrinsic information perspective, each causal mechanism (includes the  causes and effects of any state) in the system acts as an information bottleneck. This means that its cause information only exists for the system to the extent that it also specifies effect information, and vice versa. 
\vspace{-4mm}\begin{lemma}
\label{lemma_intrinsicInfo}
We define the intrinsic information learned by the expert and imitator as $\mathbb{I}_{ce}^{E}$ and $\mathbb{I}_{ce}^{\eta}$, respectively.
In the presence of confounding variables $\btheta$, the imitator learns less intrinsic information, $\mathbb{I}_{ce}^{\eta} < \mathbb{I}_{ce}^{E}$, compared to the expert. Under the assumption of perfect $\bms^t$ being extracted by the imitator, the difference $\mathbb{I}_{ce}^{\eta} - \mathbb{I}_{ce}^{E}$ can be shown to be strictly smaller than $\epsilon$ when $D_{TV}\left(p(\btheta\mid\bmH_t),\delta_{\btheta}(\btheta^0)\right) < \epsilon$, where $D_{TV}$ is the total variation distance (see Appendix~\ref{Homology}) and $\delta_{\btheta^0}(\btheta)$ is the Dirac-delta distribution, with $\btheta^0$ the true value.  
\vspace{-4mm}\end{lemma}
\begin{IEEEproof} See Appendix~\ref{lemma_intrinsicInfo_proof}.
\end{IEEEproof}
\indent Lemma~\ref{lemma_intrinsicInfo} emphasizes the significance of precise modeling of the confounding variables (known at the DT side) at the imitator to extract the relevant semantics. Moreover, we introduce the following concept, which enables the DT-enabled transmitter to generate a semantic representation that captures only the relevant and distinctive causal states with nonzero intrinsic information.
\vspace{-4mm}\begin{definition}\label{Def_StateAbs}
We define the \emph{causal-invariant state abstraction} as the $D-$dimensional embedding of the states, $\phi_i:\bms\rightarrow \mR^D$. $\phi_i$ is causal-invariant if $\forall \bms_c, \bms_e, \bms_1, \bms_2 \in \mS$, $\bma \in \mA$, $\phi_i(\bms_1)=\phi_i(\bms_2)$ if and only if the cause and effect set are same for both $\bms_1,\bms_2$ 
while satisfying $P(\bms_1\mid\bms_c,\bma)=P(\bms_2\mid\bms_c,\bma), P(\bms_e\mid\bms_1,\bma)=P(\bms_e\mid\bms_2,\bma)$.
\vspace{-4mm}\end{definition}
From Definition~\ref{Def_StateAbs} we can conclude the following: if $\phi_i(\bms_1) \!=\! \phi_i(\bms_2)$, then the intrinsic information conveyed by $\bms_1$ and $\bms_2$ is the same. In other words, from an SC perspective, state abstraction enables a unique representation for states that share comparable cause-and-effect repertoires. This leads to a reduction in the amount of semantic information conveyed compared to systems that do not employ state abstraction. The concept of state abstraction presented here differs significantly from the traditional ML definitions such as those in \cite{TomarArxiv2021} and \cite{AbelICML2018}. Next, using the intrinsic information concept, we analyze the information content provided by a state $\bms_i^t$ composed of several SCEs $U_j\in \mU$.

\vspace{-6mm}\subsection{Information Integration (via Compositionality, for Identifying Semantic Content Elements)}
\label{Info_Integ}
\vspace{-2mm}

At the level of an individual causal mechanism (state + its cause and effect mechanisms), the integration postulate \cite{OizumiPLOS2014} states that only mechanisms that specify integrated information can contribute to consciousness. Inspired by this, \emph{integrated information} for an SCM is information that is generated by the whole mechanism beyond the information generated by its parts, meaning that the mechanism is irreducible with respect to information. Similar to cause-effect information, integrated information (denoted by $\mathbb{I}_{\phi}$) is calculated as the Wasserstein distance between two probability distributions. 

To compute the quantity of information that is integrated across the $m$ parts of a system, we can partition the system into $m$ parts such as $\bmM_1^t, \bmM_2^t, \cdots, \bmM_m^t$. This partition $p_k \in \mathcal{P}_S$ (the set of all partitions of $\mS$) is defined such that $\cup_i \bmM_i^t = \bms_i^t$ and $\bmM_i^t \cap \bmM_j^t = \emptyset$. The measure of integrated information with respect to the cause and effect mechanisms can be expressed as follows:
\vspace{-2mm}\beq
\vspace{-0mm}\vspace{-0mm}\begin{array}{l}
\mbI_{\phi,c}^{p_k} = \mbI_c(\bms_i^{t-1};\bms_i^t) - \sum\limits_j \mbI_c(\bmM_j^{t-1};\bmM_j^{t}), \,\,\,
\mbI_{\phi,e}^{p_k} = \mbI_e(\bms_i^{t+1};\bms_i^t) - \sum\limits_j \mbI_e(\bmM_j^{t+1};\bmM_j^{t}).
\end{array}
\label{eq_Iphi_c_e}
\vspace{-2mm}\eeq
The integrated information for partition $p_k$ will be:
\vspace{-3mm}\beq
\mbI_{\phi}^{p_k} = \min (\mbI_{\phi,c}^{p_k} ,\mbI_{\phi,e}^{p_k} ).\label{eq_Iphi_min}
\vspace{-2mm}\eeq
$\mbI_{\phi}$ satisfies $0 \leq \mbI_{\phi} \leq \min \left({\mbI}_c\left(\bms_i^{t-1};\bms_i^t\right),\mbI_e\left(\bms_i^{t+1};\bms_i^t\right)\right)$. Thus, the SCM is reducible if at least one partition $p_k \in \mathcal{P}_S$ makes no difference to the cause or effect probability, i.e., $\mbI_{\phi}^{p_k} = 0$.
If it is zero, then the partition does not contribute any shared information to the system as a whole. Next, we define the integrated information of an SCM as given  by it irreducibility over its minimum partition $p_k \in \mathcal{P}_S$:
\vspace{-2mm}\beq
\begin{array}{l}
\mbI_{\phi} = \mbI_{\phi}^{p_k^*}, \\
\textrm{s.t.}\,\,\, p_k^* = \argmin\limits_{p_k} \frac{\mbI_{\phi}^{p_k}}{\max\limits_{p_i\in\mathcal{P}_S}\mbI_{\phi}^{p_i}}.
\end{array}
\label{eq_Iphi}
\vspace{-2mm}\eeq
The normalization above is over the maximum possible value that $\mbI_{\phi}^{p_k} $ could take for any partition. 

Having defined the integrated information measure, we consider the IL setup. The imitator relies on the probability distributions obtained from the received semantic information to measure its $\mathbb{I}_{\phi}$. To extract the maximum information, the decoding must be performed optimally using the ``true" conditional distribution, which is also a function of confounding variables. 
\vspace{-3mm}\beq
p(\bms_i^{t},\btheta\mid \bms_i^{t-1}) = p(\bmM_1^{t},\cdots,\bmM_m^{t}, \btheta\mid \bmM_1^{t-1},\cdots,\bmM_m^{t-1}).
\vspace{-2mm}\eeq
To decode $\bms_i^{t-1}$, the imitator uses a ``false" conditional distribution, $q(\bmsh_i^{t},\bthetah\mid \bms_i^{t-1})$ (false, since $\btheta$ is unknown at the imitator), and hence it is an instance of ``mismatched" decoding. $\bmsh_i^{t}$ is the received semantic information. To quantify integrated information, we specifically consider the mismatched decoding that uses the ``partitioned" probability distribution $q(\bmsh_i^{t},\bthetah\mid \bms_i^{t-1})= p_{\eta}(\btheta\mid \bmH_t)p(\bms_i^{t}\mid \bms_i^{t-1},\btheta).$ Similarly, we can define the probability distribution $q(\bmsh_i^{t},\bthetah\!\mid\! \bms_i^{t+1})$. The imitator should learn (see Section~\ref{CDL_IL}) these cause and effect probabilities to extract the semantic information (not just look to reconstruct the current state $\bms_i^t$ from $\bmy^t$). The corresponding learned integrated information, defined as $\mathbb{I}_{\phi}^{\eta}$, is called as the extrinsic information (measured using the imitator's observations, which are the received signals). Next, we quantitatively analyze the error in $\mathbb{I}_{\phi}$ between expert and imitator agents for our DT-based SC system.

\vspace{-4mm}\begin{lemma}
\label{lemma_err_semInfo}
The error between the true integrated information and the extrinsic information learned by the imitator can quantified as the bias of the learned estimator of cause and effect entropy at the imitator. Here, $\mathbb{H}$ refers to Shannon's entropy, which is a measure of uncertainty.
\vspace{-2mm}\beq
\begin{array}{l}
\mathbb{I}_{\phi,c} - \mathbb{I}_{\phi,c}^e =  \mathbb{E}_{q}\left(\mathbb{H}\left(\bms_i^{t-1}\mid \bms_i^t,\btheta\right) - \mathbb{H}\left(\bms_i^{t-1}\mid \bms_i^t,\btheta)\right)\right). \\
\mathbb{I}_{\phi,e} - \mathbb{I}_{\phi,e}^e =  \mathbb{E}_{q}\left(\mathbb{H}\left(\bms_i^{t+1}\mid \bms_i^t,\btheta\right) - \mathbb{H}\left(\bms_i^{t+1}\mid \bms_i^t,\btheta)\right)\right).
\end{array}
\vspace{-2mm}\eeq
\vspace{-7mm}\end{lemma}
\begin{IEEEproof}
    See Appendix~\ref{app_lemma_err_semInfo}.\vspace{-2mm}
\end{IEEEproof}
\indent Lemma~\ref{lemma_err_semInfo} means that the error in integrated information is the minimum of the estimator (of entropy, with respect to the mismatched probability distribution $q$) bias among cause and effect transitions. In other words, the error in semantic information at the imitator is the same as the error in the transition ($\mT$, part of $\mE$) modeling.

\vspace{-5mm}\subsection{Semantic Concepts using Exclusion }
\vspace{-2mm}

A maximally irreducible cause-effect repertoire (MICE) is specified by a subset of elements, referred to as a \emph{concept}. To find the core cause of any state $\bms_i^t$, we compute $\mbI_c(\bmM^{t-1}_i \mid \bms_i^t)$ for all possible partitions $\bmM_i$, and take the maximum among them. That is, $\mbI_c^{\textrm{max}}(\bms_i^t) = \max\limits_{\bmM_i} \mbI_c(\bmM^{t-1}_i \mid \bms_i^t)$. Similarly, we can compute the maximum $\mbI_e^{\textrm{max}}(\bms_i^t)$ in the effect direction. The partitions that represent the maximum of those values are $\bmm_c^{*} = \argmax\limits_{\bmM_i} \mbI_c(\bmM^{t-1}_i \mid \bms_i^t)$ and $\bmm_e^{*} = \argmax\limits_{\bmM_i} \mbI_e(\bmM^{t-1}_i \mid \bms_i^t)$. The core cause and effect of mechanism $\bms_i^t$ are $\bmm_c^{}$ and $\bmm_e^{}$, respectively. Together, they specify the ``what" of the concept of $\bms_i^t$. A mechanism that specifies MICE constitutes a concept, or more specifically, a core concept. A concept $\bms_i^t$ is composed of several atomic units $U_i$, referred as \emph{SCEs}. 
\vspace{-2mm}\begin{definition}
A \emph{semantic content element} can be formally defined as an atomic mechanism, with possible minimum integrated information among all partitions $p_i$.
\vspace{-3mm}\end{definition}

\subsubsection{Sub-Optimal Strategy for SCE Identification}
\label{alg_SCE}

To correctly compute \eqref{eq_Iphi_c_e}, it is necessary to know the true transition probabilities that are used in the evaluation of intrinsic information. However, since the true transition dynamics are unknown, we must rely on estimating an empirical distribution from the available data. As a result, the approach outlined here represents a suboptimal scheme for identifying the SCEs. We define the input data as a sequence of entities, each of which is represented by a $d-$dimensional value, with each entity being $\bme_i \in \mU$. To extract the entities, we can utilize the $\beta$-VAE \cite{HigginsICLR2017} which is a promising approach (the discussion of which is beyond the scope here) to extract statistically independent entities present in the data. We store the entities for further processing. Further, we define a function $f_{\phi}$ to compute the integrated information for a given subset of entities, $\bmu \subseteq \mU$. The function should take the subset of entities $\bmu$ as input and return the integrated information, $\mbI_{\phi,\bmu}$ as in \eqref{eq_Iphi_min}.
To calculate the integrated information, the procedure utilizes the analytical approach outlined in Section~\ref{Info_Integ}. First, we define a function to generate all possible partitions of a given subset of entities. The function should take the subset of entities as input and return a list of all possible partitions. Further, we iterate over all partitions and compute the integrated information for each partition as in \eqref{eq_Iphi_c_e} by calling the function $f_{\phi}$.
This returns the identified SCEs, which consist of subsets of entities with non-zero integrated information and their corresponding partitions. Among the partitions with nonzero integrated information $\mbI_{\phi,\bmu}^{p_k}$, we compute the maximally irreducible partition $\bms_i^t$. The concept as captured by $\bms_i^t$, and its constituent atomic units, $U_i$, form the SCEs.
The computation of the integrated information and generation of partitions can be computationally expensive for large datasets, so it may be necessary to optimize these functions for performance.

The amount of integrated information generated by concept $\bms^t$ is the minimum between past and future:
\vspace{-2mm}\beq
\vspace{-2mm}\mbI_{\phi}^{\textrm{max}}(\bms_i^t) = \min \left(\mbI_c^{\textrm{max}}(\bmm_c^{*}\mid \bms_i^t) ,\mbI_e^{\textrm{max}}(\bmm_e^{*} \mid \bms_i^t) \right).
\vspace{-2mm}\eeq
The set of all concepts within an SCM constitutes its conceptual structure, which can be represented in concept space. Concept space is a high-dimensional space, with one axis for each possible past and future state of the system. In this space, each concept is represented as a "star," with its coordinates given by the probability of past and future states in its cause-effect repertoire. The size of the star represents the value of its $I^{\textrm{max}}_{\phi}(\bms_i^t)$. If $I^{\textrm{max}}_\phi{(\bms_i^t)}$ is zero, the concept does not exist and if it's small, the concept exist to a minimal extent. The observed state $\bms^t$ at the DT-based expert agent comprises various semantic concepts, each designated as $\bms_i^t$. Even though previous literature, e.g., \cite{ShaoArxiv2022}, has briefly mentioned the data observed at the transmitter as composed of multiple concepts, purely as a probabilistic transformation, the proposed rigorous formulation based on theory of consciousness described in this section is the first of its kind. Up to this point, we have identified the SCEs present in the data and analyzed the information they contain. Our next step is to comprehend the associations between the different SCEs and how they collectively form a semantic concept, ultimately creating a causal graph that gives structure to the data.

\vspace{-5mm}\subsection{Defining Causal Relationships via Topological Perspective}
\vspace{-2mm}

For simplicity, we remove the dependence on $t$ in this subsection. Any concept $\bmm \subseteq \bms$ with $\mbI_{\phi}(\bmm) > 0$ specifies a candidate distinction
$d(\bmm) = (\bmm, \bmz^{*}
, \mbI_\phi(\bmm))$ within the system $\mS$ in state $\bms$. $\bmz^{*} = \{\bmz_c^{*},\bmz_e^{*}\}$ represents the maximal cause-effect pair for $\bmm$. For any state $\bms$, the causal distinctions that represent its subsets with nonzero integrated information are defined as:
\vspace{-2mm}\begin{equation}
\begin{array}{l}
\mathcal{D}^{}(\mT_{S},\bms) =  \{d(\bmm): \bmm \subseteq \bms, \mathbb{I}_{\phi}(\bmm) > 0, \bmz_c^{}(\bmm) \subseteq \bms^{\prime}_c,\bmz_e^{}(\bmm) \subseteq \bms^{\prime}_e \},
\end{array}
\vspace{-2mm}\end{equation}
where $d(\bmm)$ is the set of nonzero integrated information for $\bmm$ and $\bms^{\prime} = \{\bms_c^{\prime},\bms_e^{\prime}\}$ is the maximal cause-effect pair for the system as a whole. \emph{Causal relationships} refer to the way in which the causes and/or effects of a set of distinctions $\mathcal{D}^{}(\mT_{S},\bms)$ in a complex system overlap. In the same way that a distinction $\bmm$ identifies which units/states constitute a cause and the resulting effect, a relationship identifies which units/states correspond to which units/states among the purviews of a set of distinctions. These relationships reflect how the cause-and-effect power of its concepts is interconnected within the complex system. In short, they help to attribute \emph{meaning (semantics)} to the state $\bms$. Understanding causal relationships is crucial, as these models enable DT-based wireless systems to distill knowledge and experiences, similar to the human mind, to make reliable predictions, generalizations, and analogies. 
These qualities are essential for enabling compositional and counterfactual reasoning, active intervention in the world to test hypotheses, and the ability to articulate one's understanding to others. We now look at how to quantify the causal relationships for our system. The degree of irreducibility resulting from this binding of cause-and-effect power is measured by the irreducibility of the relationships, denoted by $(\mbI_{\phi_r} > 0)$.
In the presence of confounding variables, the learned causal distinctions by the imitator follows the relation, $\mathcal{D}(\mT_{S},\bms\mid \widehat{\btheta}) \subseteq \mathcal{D}^{}(\mT_{S},\bms\mid \btheta)$. 
As in \cite{TononiEntropy2019}, the relations between any two distinctions can be defined as follows. For any state $\bms$, consider two causal distinctions $d(\bmm_1)$ and $d(\bmm_2)$, that have a possibly non-empty overlapping atomic units as their constituents. The relation between $d(\bmm_1)$ and $d(\bmm_2)$ can be defined as the maximally irreducible subset that is common to both of them. For each candidate overlap, the intrinsic difference is assessed at the maximum information partition (MIP) for each partition, and summed together, which represents the irreducibility measure. If we consider a specific set of distinctions $\bmd \subseteq \mD(\mT_S,s)$, it is possible that there are multiple sets of causes and/or effects, denoted as $z$, such that
\vspace{-2mm}\beq
z: z \cap \{z_c^*(d),z_e^*(d)\} \neq \emptyset \, \forall d \in \bmd, \cap_{z \in \bmz}z \neq \emptyset, \abs{\bmz} > 1,
\vspace{-2mm}\eeq
with maximal overlap (called the ``faces" of the relation), $o^*(z) = \cap_{z\in \bmz} z \neq \emptyset.$
A relation $r(\bmd)$ thus consists of a set of distinctions $\bmd\subseteq \mD(\mT_{S},\bms)$, with an associated set
of faces $f(\bmd) = \{f(\bmz)\}_{\bmd}$ and irreducibility measure $\phi_r > 0$ and can be written as $w(\bmd) = \left(\bmd,f(\bmd),\phi_r\right).$ $w(\bmd)$ can be topologically represented as shown below.
\vspace{-4mm}\begin{lemma}
The structure of causes and effects in an SCM can be represented using an \emph{abstract simplical complex} \cite{HatcherCUP2002}, in which causes and effects are the vertices and the relations are simplices; relations between pairs of causes and effects are represented as 2-simplices (edges); relations between trios are represented as 3-simplices (faces) and so on.
\vspace{-4mm}\end{lemma}
The Lemma, whose proof follows directly from its definition states that a relation $r(\bmd)$ can be topologically characterized as a $\abs{\bmd}$-simplex, where $\abs{\bmd}$ is the order of a relation. Relations between two purviews are known as 2-relations, relations involving three purviews are referred to as 3-relations and so on. More details on simplex is provided in Appendix~\ref{Homology}.
The degree of a relation is defined as the number of causal distinctions present in $\bmd$. The ``faces" of the relation is same as the concept of faces in simplical homology. Further, we look at defining the irreducibility measure $\phi_r$. For any relation $w\in \mW$, the irreducibility can be defined as 
  \vspace{-2mm}\beq
\phi(\mathcal{R} \mid \mathcal{O}) = \sum\limits_i d_{\mathcal{O}}(\mS_i),
  \vspace{-2mm}\eeq
where $\mathcal{O}$ represents a candidate overlap between two causal distinctions, $\mS_i$ represents a particular partition of $\mathcal{O}$ and $d_{\mathcal{O}}(\mS_i) = D(p(\mathcal{O})\mid\mid p(\mS_i))$. Hence, $d_{\mathcal{O}}(\mS_i)$ represents the information the partition $\mS_i$ provides within a specific overlap $\mathcal{O}$.   Finally, the maximally irreducible overlap is the $\mathcal{O}$ that maximizes $\phi$:
\beq
  \vspace{-2mm}\begin{array}{l}
\mathcal{O}_{\textrm{max}}= \argmax\limits_{\mathcal{O}} \phi(\mS\mid \mathcal{O}).  \\
\phi_{R} = \phi(\mS\mid \mathcal{O}_{\textrm{max}}).
\end{array}
  \vspace{-1mm}\eeq
We now formally define semantics using the abstract simplical complex-based definition of causal relations.

\subsubsection{Topological characterization of semantics}

Semantics are represented  using the relations defined in the previous subsection, as follows:
  \vspace{-4mm}\begin{theorem}
\label{def_topos}
An abstract cell complex $\mC = \left(\mD,\prec_w, \dim\right)$ is a set $\mD$ of abstract SCEs, which are the cells here, equipped with a bounding relation $\prec_w$ and a dimension function assigning to each $\bmd\in \mD$ a non-negative integer ($\dim$), satisfying the following properties
\begin{itemize}
  \vspace{-2mm}\item if $\bmd_1\prec_w\bmd_2$ and $\bmd_2\prec_w\bmd_3$, then $\bmd_1\prec_w\bmd_3$ (called as the \emph{transitivity property})
\item if $\bmd_1\prec_w\bmd_2$, then $\dim(\bmd_1) < \dim(\bmd_2)$ (called as the \emph{monotonicity property}).
\end{itemize}
\vspace{-4mm}\end{theorem}
\begin{IEEEproof}
    \vspace{-2mm}
The proof is given in Appendix~\ref{def_topos_proof}. 
\end{IEEEproof}
Further, we look at what Theorem~\ref{def_topos} entails for the DT-based SC system. $\bmd\! \in \!\mD$ represents a causal mechanism (concept) with non-zero integrated information. The dimension of $\bmd$, denoted $\dim(\bmd)$, represents the number of distinct SCEs part of $\bmd$. The relation between two causal mechanisms $\bmd_1$ and $\bmd_2$ can be defined as the causal relationship (the overlap in the causal distinctions) between them. The usefulness of the Theorem~\ref{def_topos} for our DT-based SC system can be two-fold: Firstly, just like in natural language, where syntax refers to the grammatical forms used to express content and semantics refers to the meaning attributed to those syntactic expressions, in communication between intelligent agents, the encoded representation refers to the syntax part \cite{ChristoTWCArxiv2022}. The transmit encoders, which form a component of the semantic language used to convey meaning, can be connected to the specific problem formulation. The design of the AI architecture employed for this purpose may be constrained by the complexity of the transmitter, such as the maximum number of parameters or layers it can support. Regardless of this dependency on language, if the receiver can extract the semantic content defined using a topological construct such as a cell complex, it should be able to communicate seamlessly with multiple transmitters without requiring any protocol changes. In other words, regardless of the encoded representations used to convey the states observed at DT, the topological characterization of semantics acts as a ``bridge" \cite{CaramelloOUP2018} that unifies different semantic contents and facilitates communication between intelligent agents. The abstract cell complex structures in Theorem~\ref{def_topos} can be seen as ``universal translators" and bridges across different knowledge representations. Additionally, the topological characterization presented in Theorem~\ref{def_topos} allows a rigorous formulation of semantic metrics, such as similarity and reliability, as defined in Section~\ref{TxDesign}, compared to the current state of the art. Having defined SCEs present in the observed states, we further move on to discovering the causal structure in the data using novel AI tools. 

\vspace{-2mm}\section{Causal Discovery and Inference via Imitation Learning and model based reinforcement learning}
\label{CDL_IL}
\vspace{-1mm}

 In the context of DT-based SC, the causal dynamics model $p(\bms^{t+1}\!\mid\!\bms^t,\bma^t,\btheta)$ cannot be accessed by the imitator due to hidden confounding variables (known only to DT). Instead, the imitator can only estimate the transition distribution based on its interactions with the expert. To enable the imitator to make better control decisions, it must have the ability to develop representations of the world based on past experiences (received semantics from the DT node), which allow for generalization to novel situations. To achieve this, the transmitter design must incorporate causal discovery, state transition probability, and semantic encoding such that the QoE is maximized (i.e., expected imagined rewards $\mathbb{E}(\sum\limits_{\tau=t}^H \gamma^{\tau-t}r_{\tau})$), as discussed next.

\vspace{-6mm}\subsection{Causal Imitation Learning Problem: Transmitter Design}
\label{TxDesign}
\vspace{-2mm}

Herein, at the expert node, our objective is to throw away irrelevant state variables while learning the causal dynamics and hence design an abstract causal state description (represented using $p(\bmz^t\mid\bms^t,\bms^{t+1})$) which gets transmitted. Towards this objectives, we adopt the concept of deep end-to-end causal inference (DECI) from \cite{GeffnerArxiv2022}. 
We specifically selected DECI because other contemporary works treat causal discovery and causal inference as distinct problems. In contrast, DECI is the only approach that simultaneously addresses both problems by conducting causal discovery through the estimation of a sparse directed graph and performing causal inference by estimating specific quantities based on a given set of inputs (interventions) -- such as estimating $\bms^{t+1}$ in this particular case. Note that the objectives for causal inference in an SC system differ from those in \cite{GeffnerArxiv2022}. In contrast to \cite{GeffnerArxiv2022}, we suggest a constrained optimization approach to achieve a minimum QoE. 
DECI employs a Bayesian perspective in its causal discovery process, where the causal graph $G$ is modeled in conjunction with the observations $\bms^t,\bms^{t+1}$. First, we look at the joint distribution
\vspace{-2mm}\beq
p_{\bPsi}(\bmz^t,\bms^t,\bms^{t+1},\bma^t,\btheta,G) = p(G)p(\btheta)p(\bms^t)\pi_E(\bma^t\mid\bms^t,\btheta)p(\bms^{t+1}\mid\bms^t,\bma^t,\btheta)p(\bmz^t\mid\bms^t,\bms^{t+1}),
\vspace{-2mm}\eeq
where $\bmG$ is defined as the adjacency matrix (corresponding to the graph $\mG$), with entries $G_{j,i} \in \{0,1\}$ indicating the presence of an edge from $j\rightarrow i$. $\bPsi$ is the set of NN parameters. The causal relationships can be modeled using a non-linear additive noise model (ANM) as follows: $
\bms_i^{t+1} = \bmf_i(\bms_i^t,Pa_i) + \bmn_{i}, $
where  $\boldsymbol{n}_{i}$
is an exogenous noise variable
that is independent of all other variables. We propose a flexible NN parameterization that satisfies the graph adjacency constraints by setting
$
\bmf_i(\bms_i^{t+1}) = \boldsymbol{\zeta}_i\left(\sum\limits_j G_{j,i}\kappa_j(\bms_j^t)\right),$
where the each entry of $\boldsymbol{\zeta}_i$ and $\kappa_j$ are multilayer perceptrons (MLPs).
Our objective is to use observational data (that involves network state $\bms^t$) to fit the parameters $\bPsi$ of our non-linear ANM. After fitting the model, the posterior $p_{\bPsi}(\bmG\!\mid\!\bms^t,\bms^{t+1},\bma^t,\btheta)$ reflects our understanding of the causal structure. The graph prior $p(\bmG)$ should characterize the graph as a \emph{directed acyclic graph (DAG)}. We choose it as $p(\bmG) \propto exp(-\lambda_s\norm{\bmG}^2 -\rho\, h(\bmG)^2 -\alpha \,h(\bmG)),$
where the DAG penalty $h(\bmG) = \tr(e^{\bmG\odot \bmG}) -D$,
which is non-negative and zero only if $\bmG$ is a DAG. Variables $\alpha$ and $\rho$ represent the weights for the DAG penalty \cite{ZhengArxiv2018}, and they are  optimized during training as in \cite[Appendix B]{GeffnerArxiv2022}. The prior knowledge about graph sparseness is modeled by the term $\lambda_s\norm{\bmG}^2$, where $\lambda_s$ is a fixed known quantity. The model has two difficulties: first, the actual posterior over $\bmG$ cannot be computed efficiently, and second, the maximum likelihood approach cannot be applied to estimate the model parameters due to the presence of the latent variable $\bmG$. To address both challenges simultaneously, we employ variational inference, as suggested in previous literature \cite{BleiJASA2017}. Specifically, we introduce a variational distribution $q_{\bPsi}(\bmG)$ to approximate the intractable posterior $p_{\bPsi}(\bmG\mid\bmz^t,\bms^t,\bms^{t+1},\bma^t,\btheta)$, and utilize it to construct the evidence lower bound (ELBO) given by:
\vspace{-2mm}\beq
\begin{array}{l}
g_E(\boldsymbol{\chi},\bPsi) =  \E_{q_{\bPsi}(\bmG)}\left[\log p((\bmG)\prod\limits_{n:x^{(n)} \in \{\bmz^t,\bms^t,\bms^{t+1},\bma^t,\btheta\}} p_{\boldsymbol{\chi}}(x^{(n)}\mid \bmG)\right] + \mathbb{H}(q_{\bPsi}) \leq \log p_{\boldsymbol{\chi}}(x^{(1)},\cdots,x^{(n)}).
\end{array}
\label{eq_ELBO}
\vspace{-2mm}\eeq
Here $\boldsymbol{\chi}$ is the NN parameters for the $p_{\boldsymbol{\chi}}(x^{(n)}\mid G)$.  \eqref{eq_ELBO} can be rewritten as
\beq
\vspace{-2mm}\begin{aligned}
g_E(\boldsymbol{\chi},\bPsi) &=   \E_{q_{\bPsi}(\bmG)}\!\!\left[\!\prod\limits_{n:x^{(n)} \in \{\bmz^t,\bms^t,\bms^{t+1},\bma^t,\btheta\}} p_{\boldsymbol{\chi}}(x^{(n)}\mid \bmG)\!\right] \!-\! \textrm{KLD}\left(q_{\bPsi}(\bmG\mid \bmx)\!\mid\mid\! p(\bmG)\right) \\ & \leq \log p_{\boldsymbol{\chi}}(x^{(1)},\cdots,x^{(n)}).
\end{aligned}
\label{eq_ELBO2}
\vspace{-2mm}\eeq
The causal discovery architecture  includes a graph NN (GNN) \cite{ScarselliGNN2008} which accepts as input $\bms^t$ and propagates information
across a fully connected graph $\mG = \{\mV, \mE\}$. This graph includes vertices $v_i \in \mV$ for each $\bms_i^t$, and each pair of vertices $(v_i,v_j)$ is connected by an edge whose embedding is represented by the MLP $\bme_{i,j} = \bmf_{i,j}(\bms_i^t,\bms_j^t),$ and:
\vspace{-5mm}\beq
\begin{aligned}
p_{\bPsi}(G_{j,i}\mid\bms^t,\bms^{t+1},\bma^t,\btheta) &= \textrm{Softmax}(\bme_{i,j}\mid T),
\end{aligned}
\vspace{-3mm}\eeq
where $T$ represents the softmax temperature variable \cite{LowePMLR2022}.
DECI aims to demonstrate that the optimization of the variational objective in equation \eqref{eq_ELBO} can recover the ground truth data generation mechanism. To establish this statistical guarantee under the correct specification of DECI, which assumes the existence of $\boldsymbol{\chi}^{0}$ and $\bmG^0$ that correspond to the true data generating process and the absence of unobserved confounding variables, two technical assumptions must be satisfied. Firstly, function $\bmf_i$ must be non-invertible (since a DAG), 3rd-order differentiable, and not constant with respect to any of its inputs. Secondly, the proper noise densities must have bounded likelihood. These assumptions rule out non-identifiable ANM. Further, the ELBO maximization problem $\mP_1$ in \eqref{eq_P1} that involves expert policy computation, causal discovery and state transition dynamics can be reformulated as follows. 
\vspace{-2mm}\beq
\vspace{-2mm}\begin{array}{l}
\mP_1: \argmax\limits_{\pi_{E}(\bma^t\mid \bms^{t+1},\bms^{t},\btheta), \,\bmG,\, p(\bms^{t+1}\mid \bms^{t},\bma^{t},\btheta)}  g_E(\boldsymbol{\chi},\bPsi) \\
\mbox{subject to}\,\, \mathbb{E}(r_t)
\leq R_b
\end{array}
\label{eq_elbo_opt}
\eeq
We now explain how to obtain the reward constraint bound, denoted as $R_b$ in this context. To accomplish this, we must first define the concept of ``semantic reliability". The expert agent's goal is to select an optimal semantic representation (encoder) that accurately represents the causal structure at the transmit side. The imitator, on the other hand, aims to improve semantic effectiveness on its side to achieve a desired level of semantic reliability. \emph{Semantic reliability} is quantified by the expression, 
$
p\left(E_t\left({\bms}^t,{\widehat{\bms}}^t \right) < \delta \right) \geq 1-\epsilon,
$
where $E_t(\bms^t,{\widehat{\bms}}^t) = \norm{\bms^t-\bmsh^t}^2$ represents semantic distortion and $\epsilon$ is arbitrarily small. This metric reflects the imitator's ability to reliably reconstruct all the causal aspects in the decoded causal structure. Unlike classical reliability measures, in semantics, we can recover the actual meaning of transmitted messages even with a higher bit error rate (BER), as long as the semantic distortion remains within the set limit. This is illustrated by the choice of $\delta$ here, which depends on the concept of semantic space. The \emph{semantic space} $\mK$ is defined as an $N$-dimensional topological ball (since $\bms^t$ belongs to an abstract cell complex as proved in Theorem~\ref{def_topos}) centered at the actual state $\bms^t$, where all points inside the ball corresponding to the same semantic information. Formally, we can express this as $E_t(\bms^t,{\widehat{\bms}}^t) \leq \delta$ such that $\mbI_{\phi}(\bms^t)=\mbI_{\phi}(\bmsh^t)$, where $E_t$ is the topological distance between states $\bms^t$ and $\bmsh^t$, $\delta$ is the radius of the topological ball, and $\mbI_{\phi}$ is a mapping function that maps states to their corresponding semantic information. The topological distance here refers to the irreducibility measure quantifying the causal relationships between any two states $\bms^t,\bmsh^t$. The upper bound $R_b$ is defined by the reward obtained at $\left(\bmsh^t\right)^{*}$ which corresponds to $\left(\bmsh^t\right)^{*} = \argmin\limits_{\bmsh^t\in\mK} C_t(\bms^t,\bmsh^t)$.
The second challenge pertains to resolving the semantic encoder ($p_{\bPsi}(\bmz^t\mid (\bms^t,\bms^{t+1}))$) problem using a revised version of the information bottleneck principle \cite{BeckArxiv2023}. In constrast to \cite{BeckArxiv2023}, we leverage semantic information measures derived from IIT principles. As a result, our metrics go beyond conventional information theory approaches. Our objective is to determine the distribution $p_{\bPsi}(\bmz^t\mid (\bms^t,\bms^{t+1}))$, which converts the input signal $\bms$ into a representation $\bmz$, such that $\bmz$ discloses as little information as possible about $\bms$, while extracting the maximum amount of semantic information about $\bms$ in the output signal $\bmy$ $\left(\mathbb{I}_{\phi} \left(\bms; \bmy\right)\right)$. We reformulate $\mP_2$ in \eqref{eq_P2_init} as:
\vspace{-2mm}\beq
\vspace{-1mm}\begin{array}{l}
\mP_2: \argmax\limits_{p_{\bPsi}\left(\bmz^t\mid (\bms^t,\bms^{t+1})\right)}  \mbI_{\phi}(\bms^t; \bmy^t) \\
\mbox{s.t.}\,\, \mathbb{I}_{\phi} (\bms^t; \bmz^t)
\leq \mathbb{I}_b
\end{array}
\label{eq_P2}
\vspace{-1mm}\eeq
According to Definition~\ref{Def_StateAbs}, causally invariant states are represented using the same semantic representation $\bmz^t$, which is ensured by the constraint imposed on $\mathbb{I}_{\phi} (\bms^t; \bmz^t)$ in \eqref{eq_P2}.  Next, we discuss how to solve the above two optimization problems $\mP_1$ and $\mP_2$. 

\vspace{-2mm}\subsubsection{Proposed Solution}

\begin{figure}[t]\vspace{-5mm}
\centering{\includegraphics[width=7in,height=2.1in]{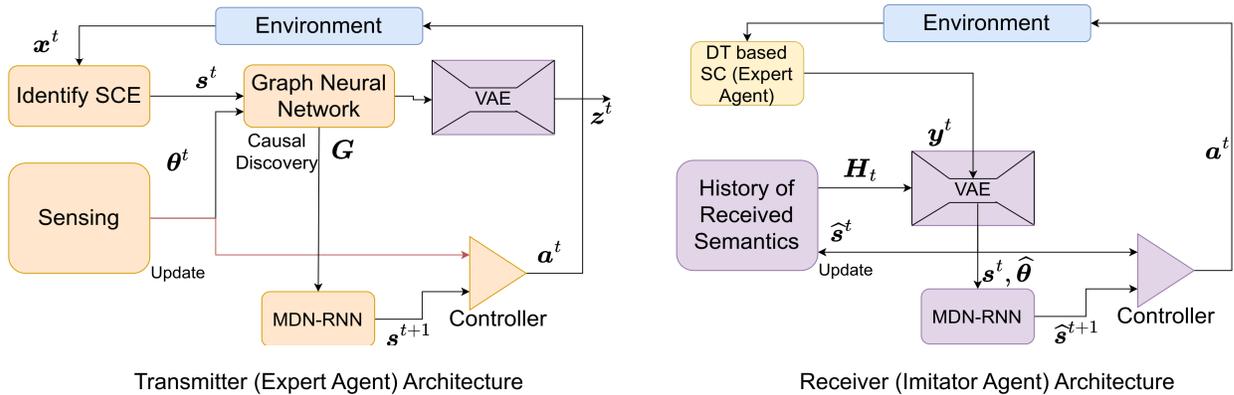}}\vspace{-5mm}
\caption{\small Generative AI components of the proposed network state model.}
\label{GenerativeAI}\vspace{-0mm}
\vspace{-0mm}
\end{figure}

The resulting bi-level optimization involving $\mP_1$ and $\mP_2$ can be alternatively solved as follows. We first write the Lagrangian corresponding to \eqref{eq_elbo_opt} and \eqref{eq_P2} as follows. 
\vspace{-2mm}\beq
\begin{array}{l}
\mL_1 = \min\limits_{\lambda_1} \max\limits_{\pi_{E}(\bma^t\mid \bms^{t+1},\bms^{t},\btheta), \,\bmG,\, p(\bms^{t+1}\mid \bms^{t},\bma^{t},\btheta)}  g_E +\lambda_1\left(R_b- \mathbb{E} (r_t)\right), \\
\mL_2 = \min\limits_{\lambda_2} \max\limits_{p_{\bPsi}\left(\bmz^t\mid (\bms^t,\bms^{t+1})\right)}  \mbI_{\phi}(\bms^t; \bmy^t) +\lambda_1\left(\mathbb{I}_b- \mathbb{I}_{\phi} (\bms^t; \bmz^t)\right).
\end{array}
\vspace{-2mm}\eeq
To solve the resulting bi-level optimization, the technique is described in Appendix~\ref{BilevelOpt}. The resulting alternating updates for outer optimization (solving $\mL_1$) can be obtained as (where $(k)$ denotes iteration):
\vspace{-2mm}\beq
\begin{aligned}\scriptsize
(\pi_{E}(\bma^t\mid \bms^{t+1},\bms^{t},\btheta))_{k+1} &= \argmax\limits_{\pi_{E}(\bma^t\mid \bms^{t+1},\bms^{t},\btheta)} \mL_1 \left(\pi_{E}(\bma^t\mid \bms^{t+1},\bms^{t},\btheta), \bmG_{(k)},p_{(k)}(\bms^{t+1}\mid \bms^{t},\bma^{t},\btheta),\lambda_{1,(k)})\right), \\
 \bmG_{(k+1)} & = \argmax\limits_{\bmG}  \mL_1 \left(\left(\pi_{E}(\bma^t\mid \bms^{t+1},\bms^{t},\btheta)\right)_{(k+1)}, \bmG,p_{(k)}(\bms^{t+1}\mid \bms^{t},\bma^{t},\btheta),\lambda_{1,(k)})\right), \\ p_{(k+1)}\left(\bms^{t+1}\mid \bms^{t},\bma^{t},\btheta\right)&= \argmax\limits_{\pi_{E}(\bma^t\mid \bms^{t+1},\bms^{t},\btheta)} \!\!\!\!\mL_1 \left(\left(\pi_{E}(\bma^t\mid \bms^{t+1},\bms^{t},\btheta)\right)_{k+1}, \bmG_{(k+1)},p(\bms^{t+1}\mid \bms^{t},\bma^{t},\btheta),\lambda_{1,(k)}\right), \\
 \lambda_{(k+1)}&= \argmin\limits_{\lambda}  \mL_1 \left(\left(\pi_{E}(\bma^t\mid \bms^{t+1},\bms^{t},\btheta)\right)_{k+1}, \bmG_{(k+1)},p_{(k+1)}(\bms^{t+1}\mid \bms^{t},\bma^{t},\btheta),\lambda_1\right).
\end{aligned}
\vspace{-3mm}\eeq
Our approach involves aggressively improving the model to maximize the ELBO based on the current semantic encoder, followed by implementing a more conservative policy and causal discovery. The detailed algorithmic process is presented in Algorithm~\ref{alg_ta_CSG} and the model architecture in Fig.~\ref{GenerativeAI}, which concludes the SC system design at the expert agent side. The algorithm convergence follows similar theoretical arguments as in \cite[Theorem 1]{GeffnerArxiv2022}. The authors therein demonstrate that maximizing the ELBO in~\eqref{eq_ELBO} can recover both the ground truth data generating process, $p(\bms^t,\bms^{t+1},\btheta,\bmz^t,\bma^t; \bmG^0)$, and the true causal graph, $\bmG^0$, in the infinite data limit. Apart from the ANM model, the convergence guarantees are under the assumption that 1) there are no latent confounders (which is true from the perspective of expert agent) and 2) the log-likelihood is regular, which means that $\mathbb{E}_p\left[\abs{\log p_{\boldsymbol \chi}(\bms^t,\bms^{t+1},\bmz^t,\bma^t\mid\bmG)}\right] < \infty$. We next look at the causal IL problem for the receiver design. 
\setlength{\textfloatsep}{0pt}
\begin{algorithm}[t]\scriptsize
\caption{Proposed Bi-level Optimization for Causal Discovery and Semantic Encoder}\label{alg_ta_CSG}
 \textbf{Given:} $p(\bmG), p(\bms^0)$ \vspace{-1mm}\\ 
  \textbf{Define:} $\mE = \left(\bmG,p(\bms^{t+1}\mid\bms^t,\btheta,\bma^t),\pi_E(\bma^t\mid\bms^t,\btheta)\right)$ \vspace{-1mm}\\ 
\textbf{Initialize:} \mbox{Policy} $\pi_E$,\,\,\mbox{causal graph} $\bmG\sim p(\bmG)$, \mbox{data buffer} $\mD = \{\}$. 
\begin{algorithmic}[1]
\vspace{-2mm}\For{ ($t=0,1,2,\dots$)}
\vspace{-1mm}\State \hspace{0cm} Collect data $\mD_t = (\bms^t,\btheta)$ by executing $\pi_E$ in the environment.
\vspace{-1mm}\State \hspace{0cm} Build local (policy-specific) dynamics model:
$\left[\mEh_{t+1},\widehat{\lambda}_{t+1}\right] = \argmin\limits_{\lambda} \max\limits_{\mE} g_E+\lambda (R_b-\mathbb{E}(r_t))$.
\vspace{-1mm}\State \hspace{0cm} Improve policy: $\left(\pi_{E}\right)_{k+1}  = \left(\pi_{E}\right)_{k} + \alpha_k \nabla_{\pi_{E}}g_E(\pi_{\eta}, \mEh^{\pi}_{k+1}).$ with a conservative algorithm like policy search with natural gradient (NPG) \cite{RajeswaranNIPS2017}. 
\vspace{-4mm}\State \hspace{0cm} Compute the encoder distribution, given the causal graph and state transitions as $p(\bmz^t\mid\bms^t,\bms^{t+1},\bmG)$ using VAE.\vspace{-1mm}
\vspace{-0mm}\State \hspace{0cm} Simulate the receiver part using Algorithm~\ref{alg_2}.
\vspace{-1mm}\State \hspace{0cm} Evaluate semantic effectiveness metric $C_t(\bms^t,\bmsh^t)$. \vspace{-1mm}
\EndFor
\end{algorithmic}
\label{algo1}  
\vspace{-1mm}\end{algorithm}
\vspace{-4mm}

\vspace{-2mm}\subsection{Causal Imitation Learning Problem: Receiver Design}\vspace{-1mm}

\vspace{-2mm}
The distribution of the trajectory $\tau$ of the expert's interactions with the environment can be written as: 
\vspace{-2mm}\beq
P_{E}(\tau\mid \btheta) = p(\bms^0\mid \btheta)\prod\limits_{t=0}^{H}p(\bms^{t+1}\mid\bms^t,\bma^t,\btheta)\pi_E(\bma^t\mid\bms^t,\btheta)
\vspace{-3mm}\eeq
Since the imitator does not have access to the confounding variables $\btheta$, it must learn the distribution over trajectories based on the state transitions history that is available to it. This history is constructed from the imitator's interactions with the environment. The imitator can then use this learned distribution to guide its own actions and improve its performance on the task. 
\vspace{0mm}\beq
\vspace{-1mm}\begin{array}{l}
P_{\eta}(\tau\mid \btheta) =  p(\bmsh^0\mid \btheta)\prod\limits_{t=0}^{H}\underbrace{p(\bmsh^{t+1}\mid\bmsh^t,\bma^t,\btheta)}_{\textrm{causal dynamics model}}\underbrace{\pi_{\eta}(\bma^t\mid \bmsh_0,\bma_0,\cdots, \bmsh^t)}_{\textrm{imitator policy}}.
\end{array}
\vspace{-1mm}\eeq
The goal of IL here is for the imitator to compute the policy such that the average reward (which is essentially QoE here) is maximized. The average is taken across all possible trajectories and the confounding variables, i.e.,
$
\mathbb{E}_{\btheta \sim P_{}(\btheta\mid \tau)}\mathbb{E}_{\tau\sim P(\tau\mid \btheta)} \left[\mR(\bmsh^t,\bma^t;\btheta)\right].$
The resulting policy learned at the imitator is $\pi_\eta(\bma^t\mid \bmsh^t,\widehat{\btheta})$.
However, this approach assumes that the imitator is ideal, i.e., it has infinite memory capacity and perceives the state and actions in the same way as indicated by the expert. Specifically, the optimal policy is often a deterministic function of the environment $\mE = (\mT,\mR,\btheta)$ such that if the imitator is able to identify the environment, it has all the necessary information to determine the optimal policy. The imitator reflects its initial uncertainty about the environment through a prior distribution $P(\mE \in \cdot)$. As the history unfolds, the imitator's current knowledge of the environment can be represented by the posterior probabilities $P(\mE \in \cdot \mid \bmH_t)$. $\bmH_t$ represents the history of state-action pairs available at the imitator node until time $t$. The total number of bits needed to identify the environment is $\mathbb{H}(\mE)$. Assuming the reward function is also unknown to the imitator (note that the expert agent evaluates the semantic effectiveness, which is the intrinsic reward), the total number bits required to learn the environment is
\vspace{-3mm}\beq
\mathbb{H}(\mE) = \mathbb{H}(\mT) + \mathbb{H}(\mR). 
\vspace{-2mm}\eeq
\indent In other words, the DT-based SC system may need to invest $\mathbb{H}(\mE)$ bits for the initial training. However, in practice, the imitator can only acquire a learning target $\mEh$, which is a stochastic function of the actual environment. This means that due to communication and computational limitations, the amount of information acquired by the agent about $\mE$ is only $\mathbb{I}(\mE;\mEh)$, which is derived from the semantic information received (states here). This can be quantified as only a subset of history being stored at the agent due to memory limitations, i.e., $P(\mE \in \cdot \mid \bmH_t^{t-K})$. Here, $\bmH_t^{t-K}$ represents the history of state transitions corresponding to the most recent $K$ time instants. The states perceived by the imitator can be erroneous due to the presence of a wireless channel between the expert and the imitator. Thus, we can consider that the imitator only knows an estimate $\bmHh_t$ and, hence, the environment $P(\mEh \in \cdot \mid \bmHh_t)$. This limits the space over which the learned target environment lies. As a result, aiming for $\mEh$ incurs a bounded degree of performance, $\mathbb{E}[\bar{V}^{*}-\bar{V}^{\pi_{\mEh}}]$. A natural measure of distortion here is the expected squared regret between the optimal and target policies:
\vspace{-2mm}\beq
\begin{aligned}
d(\mE,\mEh\mid \bmH_t) &=  
\mathbb{E}\left[\mathbb{E}_{\pi_{\mEh}}\left(\sum\limits_{t^{\prime}=t}^{H}\bar{V}^{*}(\bmH_t)-\bar{V}^{\pi_{\mEh}}(\bmH_t)\right)^2\mid \mE, \bmH_t\right].
\end{aligned}
\label{eq_avg_regret}
\vspace{-3mm}\eeq
\indent We now look at how to estimate the different components of $\mEh$. The receiver begins by extracting the semantics $\bms^t$ from the received signal $\bmy^t$, which are modeled using the channel probability distribution $p(\bmy^t\mid\bmz^t).$ The resulting $\bmsh^t$ is then stored in the observation history. 
Using this history, the receiver obtains an estimate of the confounding variables $\bthetah^t$ that helped the expert agent to analyze the optimal policy. 
The imitator then uses both the confounding variable information and the extracted state to derive the policy $\pi_{\eta}(\bma^t\mid \bmsh^{t},\bthetah^t)$. We now look at the details of the proposed semantic decoder and imitator policy design at the receiver. First, we define the loss function that captures the difference in learned state transition dynamics compared to that at the expert node.
\vspace{-3mm}\begin{definition}
 Given the inferred model at the imitator $\mEh$ and
the original sampling distribution $\mu(\bms, \bma)$, the \emph{model approximation loss} is defined as follows.
\vspace{-2mm}\beq
l(\mEh,\mu) = \mathbb{E}_{(\bms,\bma)\sim\mu} \left[\textrm{KLD}\left((P(\cdot\mid\bms,\bma)\mid\mid\widehat{P}(\cdot\mid\bms,\bma)\right)\right].\label{eq_loss}
\vspace{-2mm}\eeq
\end{definition}
\eqref{eq_loss} is used to quantify the semantic reconstruction loss on the imitator's end. As explained next, we aim to address the issue of determining the state transition probability and imitator policy.

\subsubsection{Variational Inference Framework for Receiver Design}

For simplicity, we reuse the same notation $\bPsi$ from Section~\ref{TxDesign} for the NN parameters here. Our proposed method here
learns confounding variables by maximizing an information theoretic objective defined as follows
\vspace{-2mm}\beq
\begin{aligned}
F(\bPsi) & = \mathbb{I}(\bms^t;\btheta\mid \bmH_t) + \mathbb{H}\left[\bmA \mid \bms^t;\bmH_t\right] - \mathbb{I}\left(\bmA; \btheta \mid \bms^t;\bmH_t\right)  \\
&= \left(\mathbb{H}[\btheta\mid\bmH_t] - \mathbb{H}[\btheta \mid \bms^t;\bmH_t]\right) + \mathbb{H}\left[\bmA \mid \bms^t;\bmH_t\right]   - ( \mathbb{H}\left[\bmA \mid  \bms^t;\bmH_t\right] -\mathbb{H}\left[\bmA \mid \bms^t, \btheta;\bmH_t\right]) \\
&= \mathbb{H}\left[\btheta\mid \bmH_t\right] -  \mathbb{H}\left[\btheta \mid \bms^t;\bmH_t\right] +  \mathbb{H}\left[\bmA \mid \bms^t,\btheta;\bmH_t\right].
\label{eq_vlb_confounding}
\end{aligned}
\vspace{-2mm}\eeq
We particularly derived $F(\bPsi)$ (where $\bPsi$ captures the NN parameters) for the following reasons. The first term
encourages our prior distribution over $p(\btheta\mid \bmH_t)$ to have high entropy, which implies that the imitator ideally prefer to obtain maximum information about $\btheta$.  The second term suggests that it should be
easy to infer the confounding variable $\btheta$ from the current state and history. The third term suggests that each learned target should act as
randomly as possible (random actions), which we achieve by using a maximum entropy policy to represent each confounder. As we cannot integrate over all states and confounders to compute $p(\btheta \mid \bms^t,\bmH_t)$ exactly, we approximate this
posterior with a learned discriminator $q_{\bphi}(\btheta \mid \bms^t,\bmH_t)$. By using Jensen’s Inequality, replacing $p(\btheta \mid \bms^t,\bmH_t)$
with $q_{\bphi}(\btheta \mid \bms^t,\bmH_t)$ gives us a variational lower bound $G(\bphi,\bpsi)$ on our objective $F(\bphi)$. Hence, we rewrite \eqref{eq_vlb_confounding}
\vspace{-2mm}\beq
\begin{aligned}
F(\Psi) &\geq \mathbb{H}\left[\bmA \mid \bms^t,\btheta;\bmH_t\right]  +  \mathbb{E}_{\btheta \sim p(\btheta)}[\log q_{\bphi}(\btheta \mid \bms^t;\bmH_t) -  \log p(\btheta\mid \bmH_t)] = G(\bphi,\bpsi).
\end{aligned}\label{eq_ELBO_IT}
\vspace{-2mm}\eeq
Our goal here is to maximize \eqref{eq_ELBO_IT} while simultaneously reducing the average regret (that captures the performance shortfall associated with the suboptimal learning target $\mEh$) in \eqref{eq_avg_regret}. To incorporate the latter aspect, we look at the principle of information-directed
sampling (IDS)
 \cite{ArumugamNIPS2021}. IDS is an abstract objective for sequential decision-making agents where, at each time
period, an agent computes a policy based on the current history $\bmH_t$ that minimizes the following ratio of average regret to information:
\vspace{-2mm}\beq
\begin{aligned}
J(\pi_{\eta},\mEh) & =  \frac{\mathbb{E}_{\pi}\left[d(\mE,\mEh\mid \bmH_t)\right]}{\mathbb{I}(\mEh;\bma^t,\bms_{t+1}\mid \bmH_t)}.\,\,\, 
\end{aligned}
\vspace{-2mm}\eeq
IDS is particularly relevant for CSC since it has the dual objective of (a) maximizing both the QoE by narrowing the gap between its performance and (b) that of an expert policy and enhancing the understanding of the environment $\mE$ by utilizing the current extracted semantic state and its history.
Unlike in \cite{ArumugamNIPS2021}, we propose to replace the IDS via 
\vspace{-2mm}\beq
\begin{aligned}
\left[\pi^{\mEh}_{\eta},\bthetah,\widehat{p}(\bms^t\mid\bmy^t)\right] =& \argmin\limits_{\pi_{\eta},\,\btheta,\,p(\bms^t\mid \bmy^t)} \frac{\mathbb{E}_{\pi}\left[d(\mE,\mEh\mid \bmH_t)\right]}{\mathbb{E}_{\bmy,\bms}G(\bphi,\bpsi)} \\
&\textrm{s.t.}\,\, 
p\left(E_t\left({\bms}^t,{\widehat{\bms}}^t \right) < \delta \right) \geq 1-\epsilon,
\end{aligned}
\label{eq_IDS_mod}
\vspace{-2mm}\eeq
where the constraint represents the semantic reliability measure defined earlier. For the denominator term, the expectation is over $p(\bmy^t)p(\bms^t\mid\bmy^t)$, where $p(\bms^t\mid\bmy^t)$ is the semantic decoder distribution. 
Next, we look at how to solve \eqref{eq_IDS_mod} using MBRL. 

\vspace{0mm}\subsubsection{Model Based RL as a Bi-level Optimization}
\vspace{-2mm}

MBRL is formulated as a bi-level optimization in order to capture the interactions between model and policy learning. 
Further, we look at the optimization problem for the transition model learning and the imitator policy computation. 
\vspace{-3mm}\beq
\mP_3: \underbrace{\max\limits_{\pi_{\eta}} J(\pi_{\eta},\mEh)}_{\mbox{policy-learner}}, \,\,\,\,\,\,\underbrace{\min\limits_{\mE}l\left(\mEh,\mu_{\mE}^{\pi_{\eta}}\right) \mbox{subject to}\, c(\mEh) \leq C}_{\mbox{model-learner}} 
\label{eq_bilevel_rx}
\vspace{-2mm}\eeq
We use $\mu_{\mE}^{\pi_{\eta}} = \frac{1}{T}\sum_{t=0}^T P(\bms^t = s,\bma^t=a)$ to denote the
average state visitation distribution. $c(\mEh) \leq C$ captures the resource constraints at the imitator. The MBRL can be viewed as a two-level optimization process here, and the technique to solve it is described in Appendix~\ref{BilevelOpt}. At the outer level, the objective is to optimize the policy $\pi_{\eta}$ to achieve the best possible performance within the learned model. At the inner level, an optimization problem is solved to minimize the prediction error for $\mE$ under the induced state distribution of the policy. This is a bi-level optimization because each objective depends on the parameters of both the problems discussed above. The formulation above decomposes MBRL into policy learning and generative model learning components, emphasizing that they are interdependent and must be addressed together for success. The technique to solve $\mP_3$ is discussed in Appendix~\ref{BilevelOpt} and detailed steps are described in Algorithm~\ref{alg_2}. Next, we look at the NN architecture that we adopt for various learned components in $\mP_3$.



\vspace{-8mm}\subsection{Generative AI Architecture for ``Network State Model"}
\vspace{-2mm}

The imitator model considered here is based on the concept of world models proposed in generative AI \cite{DavidNIPS2018} and is inspired by our cognitive system. Generative AI is especially advantageous for DT-based SC systems for the following reasons. First, it helps to improve the amount of semantic information acquired by the receiver through accurate modeling of the physical world with the help of DT. Second, as the information to be transmitted becomes more complex, such as 2-D images and videos, and eventually 3-D holograms or even higher-dimensional objects in future connected intelligence systems, generative AI can help the receiver generate more information with less semantics transmitted, compared to classical syntactic communication. It comprises three components that work together to process received semantic information and make decisions based on past experiences. The first component is the \emph{visual sensory component}, which VAE represents. This component is responsible for decoding the received information received by the agent into a respective network state $\bmsh^t$. This decoder distribution is represented by $p(\bms^t\mid\bmy^t,\bmH_t)$. The second component is the \emph{memory component}, represented by mixture-density network (MDN) combined with a recurrent NN (MDN-RNN). It predicts future states based on historical information, allowing the imitator to anticipate and prepare for potential future events. To anticipate the system's future states, we use a RNN as a predictive model of future $\bms^{t+1}$ vectors. However, since many complex environments are stochastic, we train our RNN to output a probability density function $p(\bms^{t+1})$ instead of a deterministic prediction of $\bms^{t+1}$. This allows us to capture the uncertainty inherent in the environment and make more informed decisions based on possible future states. In our approach, we approximate $p(\bms^{t+1})$ as a mixture of Gaussian distributions. We then train the memory component (MDN-RNN) to output the probability distribution of the next latent vector $\bms^{t+1}$, based on its current and past information. To be more specific, the RNN, which has $N_h$ hidden units, models $P(\bms^{t+1} \mid \bma^t, \bms^t, \bmh^t)$, where $\bmh^t$ is the hidden state of the RNN at time step $t$. We can adjust a temperature parameter $T$ during sampling to control model uncertainty, as in \cite{DavidNIPS2018}. We have found that adjusting $T$ is useful for training our controller later. The third and final component is the decision-making component, which is the \emph{controller}. It makes decisions based solely on the representations created by the vision and memory components, enabling the agent to take appropriate actions based on past experiences and current situations. At the time $t$, the imitator takes action $\bma^t \in \mR^{N_a}$, where $N_a$ is the dimension of the action space. The proposed imitator model has been named the \emph{``network state model"}, which shares similarities with the concept of world models in AI. However, in contrast to traditional world models in AI, the various generative AI components in the network state model (that mimic the physical environment semantics) are optimized to achieve semantic effectiveness (QoE) as close to 1 as possible, which is critical for a DT-based SC system. The training procedure to optimize the generative AI parameters is detailed in Algorithm~\ref{alg_2}. 
\setlength{\textfloatsep}{0pt}
\begin{algorithm}[t]\scriptsize
\caption{Proposed Bi-level Optimization for Receiver Design}\label{alg_2}
 \textbf{Given:} $\bmy$ \vspace{-1mm}\\ 
\textbf{Initialize:} Sample $\btheta \sim p(\btheta)$, and $\bms^t \sim p(\bms^t)$.  Collect $10,000$ rollouts from a random policy 
\vspace{-1mm}\begin{algorithmic}[1]
\vspace{-1mm}\For{ ($t=0,1,2,\dots$)}
\vspace{-1mm}\State \hspace{0cm} Train VAE to decode $\bmy^t$ into $\bmsh^t \in \mR^N$
\vspace{-1mm}\State \hspace{0cm} Update the history of observations with $\bmsh^t$.
\vspace{-1mm}\State \hspace{0cm} Train MDN-RNN to model $p(\btheta\mid\bmH_t,\bmsh^t)$. Sample the point estimate $\bthetah \sim \argmax p(\btheta\mid\bmH_t,\bmsh^t)$ as its mode. 
\vspace{-1mm}\State \hspace{0cm} Train the transition model $p(\bmsh^{t+1}\mid \bmsh^{t},\bthetah,\bmH_t)$ using RNN. 
\vspace{-1mm}\State \hspace{0cm} Evolve controller to maximize the expected cumulative reward.\vspace{-1mm}
\EndFor
\end{algorithmic}
\label{algo1}  
\vspace{-1mm}\end{algorithm}
\vspace{-1mm}
Next, we assess the performance shortfall associated with the suboptimal environment (and hence a suboptimal policy compared to expert agent) learned at the imitator. 
\vspace{-3mm}\begin{theorem}
\vspace{-2mm}\label{theorem1}
(Global performance of equilibrium pair)
Consider a pair of policy and environment model at the imitator, $(\pi_{\eta},\mE)$, such
that simultaneously
\vspace{-3mm}\beq
D_{TV}(P(\cdot\mid\bms,\bma,\bthetah),\widehat{P}(\cdot\mid\bms,\bma,\bthetah)) \leq \epsilon_{\mE}, \forall \bms, \bma, \bthetah
\vspace{-2mm}\eeq
and for simplicity, we assume that the extrinsic reward is always zero and the semantic effectiveness (intrinsic reward) is bounded, such that $\mR(\bms) \leq R_{\textrm{max}}, \forall \bms \in \mS$. Also, assume that the information learned about the environment at the imitator is incremental over time, i.e. $\mathbb{I}(\mEh;\bma^t,\bms^{t+1}\mid \bmH_t) = \mathbb{I}_0t,$ where $\mathbb{I}_0$ is information learned during the first communication instance. Then, we show that
\vspace{-3mm}\beq
\abs{J(\pi_{\eta},\mE) - J(\pi_{\eta},\mEh)} \leq \frac{2\gamma\epsilon_{\mE}R_{\textrm{max}}}{(1-\gamma)^2\mathbb{I}_0t}\,\, \forall \pi_\eta
\vspace{-3mm}\eeq
\vspace{-5mm}\end{theorem}\vspace{-4mm}
\begin{IEEEproof}
    See Appendix~\ref{theorem1_proof}.
\end{IEEEproof} 
\indent Theorem~\ref{theorem1} indicates that, with time, the discrepancy between the IDS objective and a hypothetical scenario where the imitator has perfect knowledge of $\mE$ decreases to zero. This means that as communication between the expert and the imitator continues, the imitator can accurately replicate the expert's policies.
\vspace{-3mm}\begin{theorem}
\label{err_gap}
Suppose we have policy-model pair $(\pi_{\eta},\mE)$ such that the
following conditions hold simultaneously:
\vspace{-2mm}\beq
l(\mEh,\mu^{\pi,t}_{\mE}) \leq \epsilon_{\mE} \,\,\forall t\,\, \mbox{and}\,\, J(\pi,\mEh) \geq \sup\limits_{\pi^{\prime}}J(\pi^{\prime},\mEh) -\epsilon_{\pi}.
\vspace{-2mm}\eeq
Let $\pi^{*}$ be an optimal policy so that $J(\pi^{*},\mE) \geq J(\pi^{\prime},\mE) \,\,\forall \pi^{\prime}$. Then, at time $t$, the performance of the imitator (evaluated using QoE) will be bounded, as follows:
\vspace{-2mm}\beq
J(\pi^{*},\mE) - J(\pi,\mE)  \leq \frac{2R_{\textrm{max}}}{(1-\gamma)\mathbb{I}_0t}D_{TV}(\mu_{\mE}^{\pi^{*}},\mu_{\mEh}^{\pi^{*}}) + \frac{2\gamma \sqrt{\epsilon_{\mE}}R_{\textrm{max}}}{(1-\gamma)^2\mathbb{I}_0t}.
\vspace{-2mm}\eeq
\end{theorem}
\begin{IEEEproof}
 See Appendix~\ref{err_gap_proof}.  
\end{IEEEproof} 
\indent Theorem~\ref{err_gap} shows that, following a finite number of communication instances, if $\mathbb{I}_0t$ is a finite value, the discrepancy in transition probability modeling (denoted by $D_{TV}(\mu_{\mE}^{\pi^{*}},\mu_{\mEh}^{\pi^{*}})$) dominates the error in the IDS objective with regards to the expert and imitator nodes. Therefore, accurately modeling the physical environment, specifically the network state transitions in our DT-based system, is crucial to ensure that the imitator's policy closely matches the optimal policy generated by the expert. As we observed in Theorem~\ref{theorem1}, using a generative AI architecture at the imitator can help to close the gap between the physical models of the expert and imitator.

\vspace{-2mm}\section{Simulation Results and Analysis}
\vspace{-2mm}

To validate the effectiveness of our CSC approach, we must first confirm that our learning methods can accurately identify causal relationships in the world model. We must also ensure that our objective functions and semantic metrics facilitate this process and speed up the discovery of causal relationships. 
We begin with a dataset $\mX = \{\bmx_s\}_{s=1}^S$ of $S$ samples where each sample $\bmx_s$ consists
of $N$ stationary time-series $\bmx_s = \{\bmx_{s,1},\bmx_{s,2},\cdots,\bmx_{s,{N}}\}$ across time-steps $t = \{1, \cdots, T\}$. This dataset represent the states observed at the expert agent. We denote the
$t$-th time-step of the $i$-th time-series of $\bmx_s$ as $\bmx_{s,i}^t \subset \mR^D$. We consider an SCM captured by an associated DAG $\mG_s^{1:T} = \{\mV_s^{1:T},\mE_{s}^{1:T}\}$. underlying the generative process
of each sample. The SCM's endogenous (observed)
variables are vertices $v_{s,i}^{t}\in \mV_s^{1:T}$ for each time-series $i$ and each time-step $t$.  Every set of incoming
edges to an endogenous variable defines inputs to a deterministic function $g^t_{s,i}$ which determines that
variable’s value. Using DECI, our learned NN is assumed to follow the generation of $\bmx_s^t$ as $
\bmx_s^{t+1} = \bmg_s(\bmx_s^{\leq t},\mG_s, \btheta,\bma^t) + \bmv_s^{t+1}.$
$\bmx_s^t$ is the network state (physical environment) observed by the DT. 
The edges are defined by ordered pairs of vertices $\mE_s^{1:T} = \{(v_{s,i}^{t},v_{s,j}^{t^{\prime}})\}$. Here, we consider $t^{\prime} = t-1$. and $\bmg_{s,i}^t =\bmg_{s,i}^{t^{\prime}}$. In other words, our causal graphs and process dynamics are invariant across time for each time series sample and can vary across different samples. Our expert agent NNs must model $\bmg_s$ via the transition probability and the causal graph $\mG_s^{}$. An encoded version $\bmz^t$ is extracted using a VAE, which gets communicated to the imitator agent. At the receiver, the objective is to recover the causal graph accurately and regenerate the states. The actions at the receiver belong to a discrete set (of cardinality $N_a$), with a random chosen probability. The entire state space is divided into multiple nonoverlapping subsets, where for each subset there is a unique action probability, which is a discretized Gaussian distribution centered around the mean $\widetilde{\bma}$, where the mean is distinct for different subsets. We test the CSC system on three datasets: two fully-observed physics simulations (Kuramoto and Particles)
and the Netsim dataset of simulated fMRI data that originally appeared in \cite{kipf2018neural} and was utilized for causal discovery in \cite{LowePMLR2022}.
\begin{figure}[t]
\begin{subfigure}{0.55\textwidth}
\centering{\includegraphics[width=4in,height=2in]{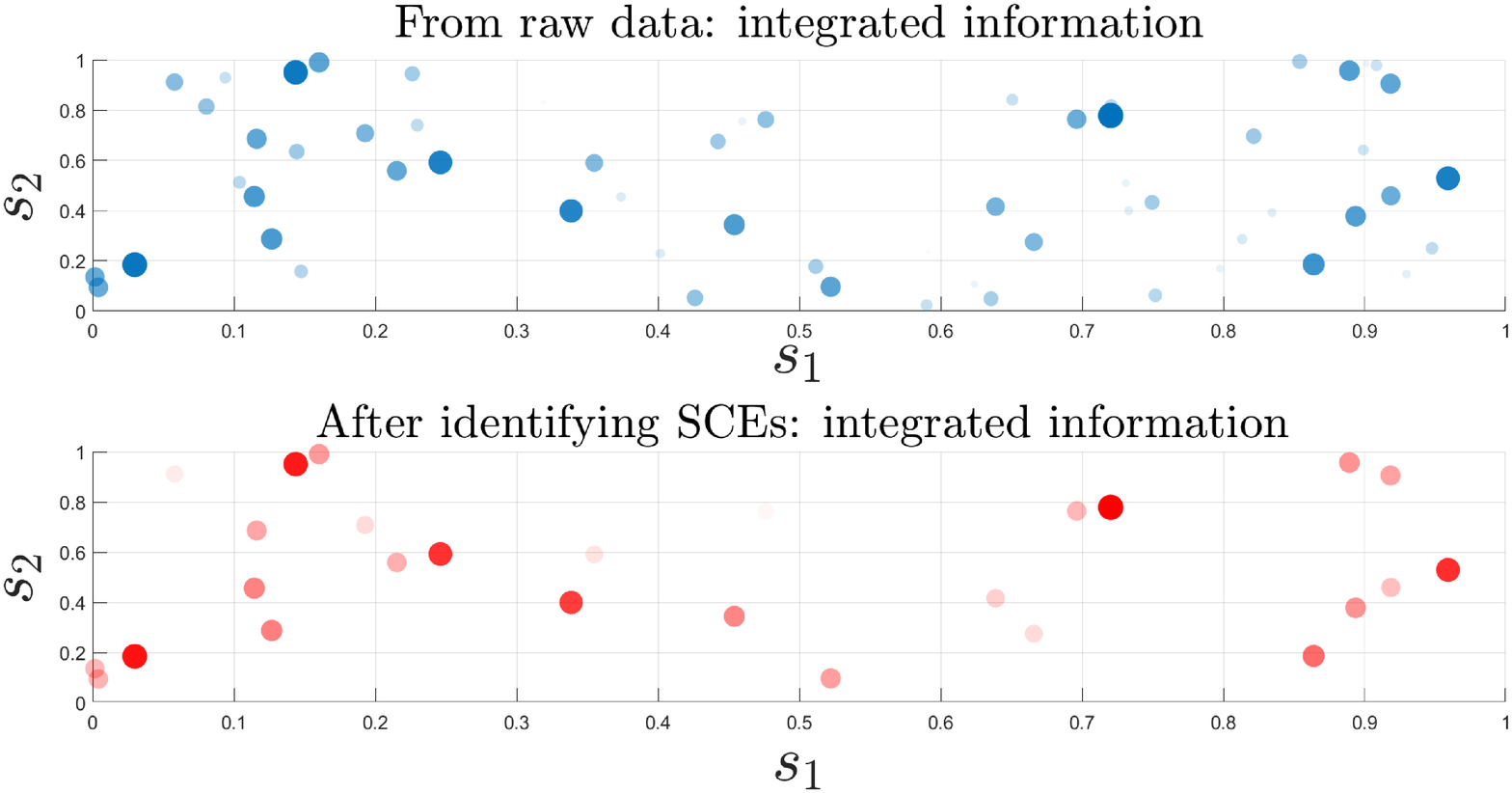}}\vspace{-3mm}
\caption{}
\label{FigIntegInfo}\vspace{-3mm}
\vspace{-2mm}
\end{subfigure}
\begin{subfigure}{0.45\textwidth}
\centering{\includegraphics[width=3.1in,height=2in]{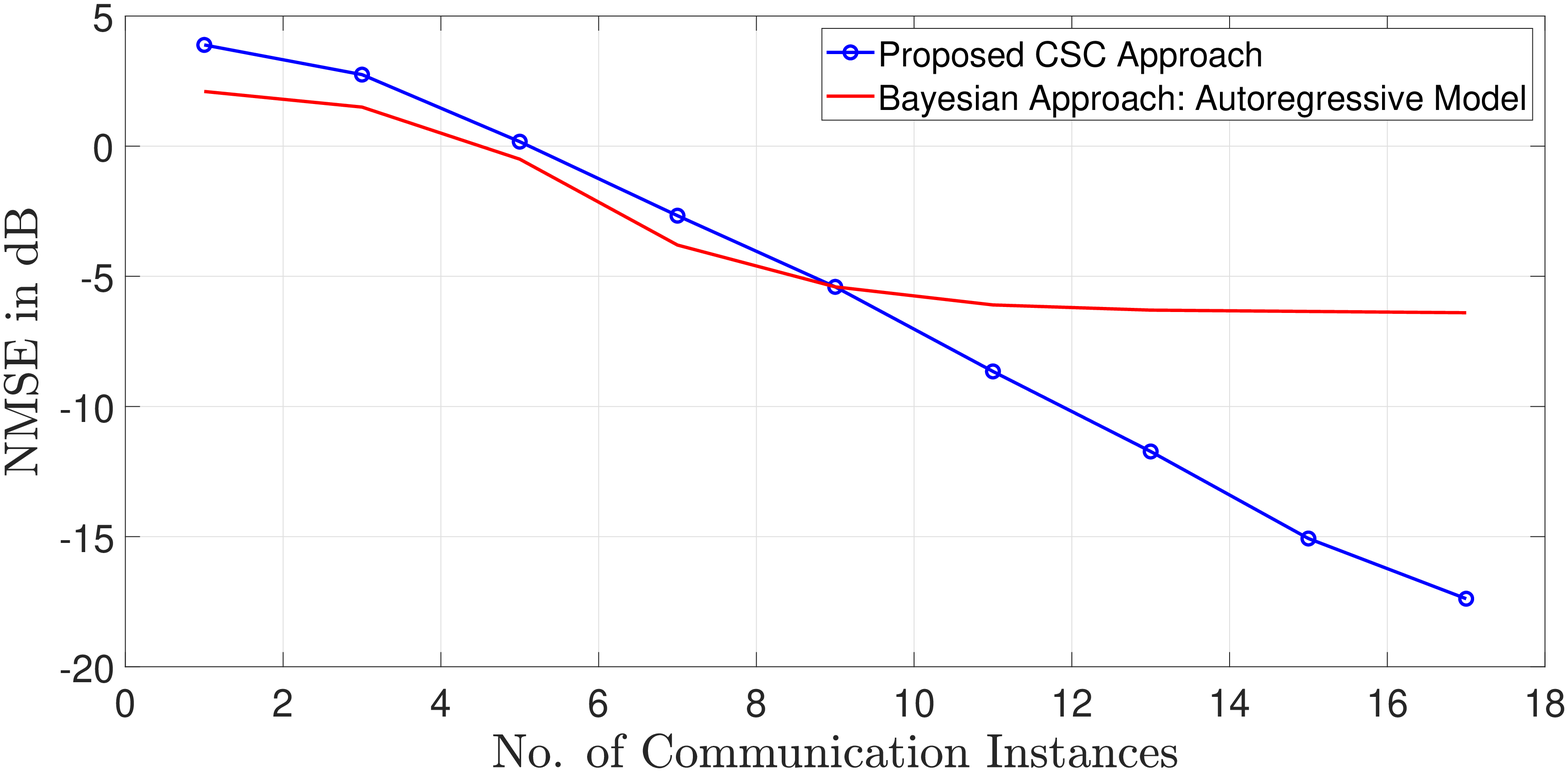}}\vspace{-3mm}
\caption{}
\label{FigConfoundingError}\vspace{-3mm}
\vspace{-2mm}
\end{subfigure}\vspace{-1mm}
\caption{\small (a) Integrated information of different subset of 
 states $\bms_i^t$. The brightness of the circles represents the magnitude of the integrated information, which is shown in a two-dimensional format. (b) Error in confounding variable estimate between expert and imitator agent.}
\vspace{-2mm}\end{figure}
Throughout the experiments, we compare the performance
of the proposed CSC framework to the two following benchmarks.  The first is a classical wireless system that directly transmits the received state information (after encoding using a VAE) without any semantic extraction. The second baseline is an SC system that is causality unaware (and hence non-generalizable) \cite{XieTSP2021}. Herein, the SCEs are identified using our proposed approach and further encoded/decoded using transformer modules as in \cite{XieTSP2021} without any causal structure extraction. 
\begin{figure}[t]
\begin{subfigure}{0.52\textwidth}
\centering{\includegraphics[width=3.1in,height=1.5in]{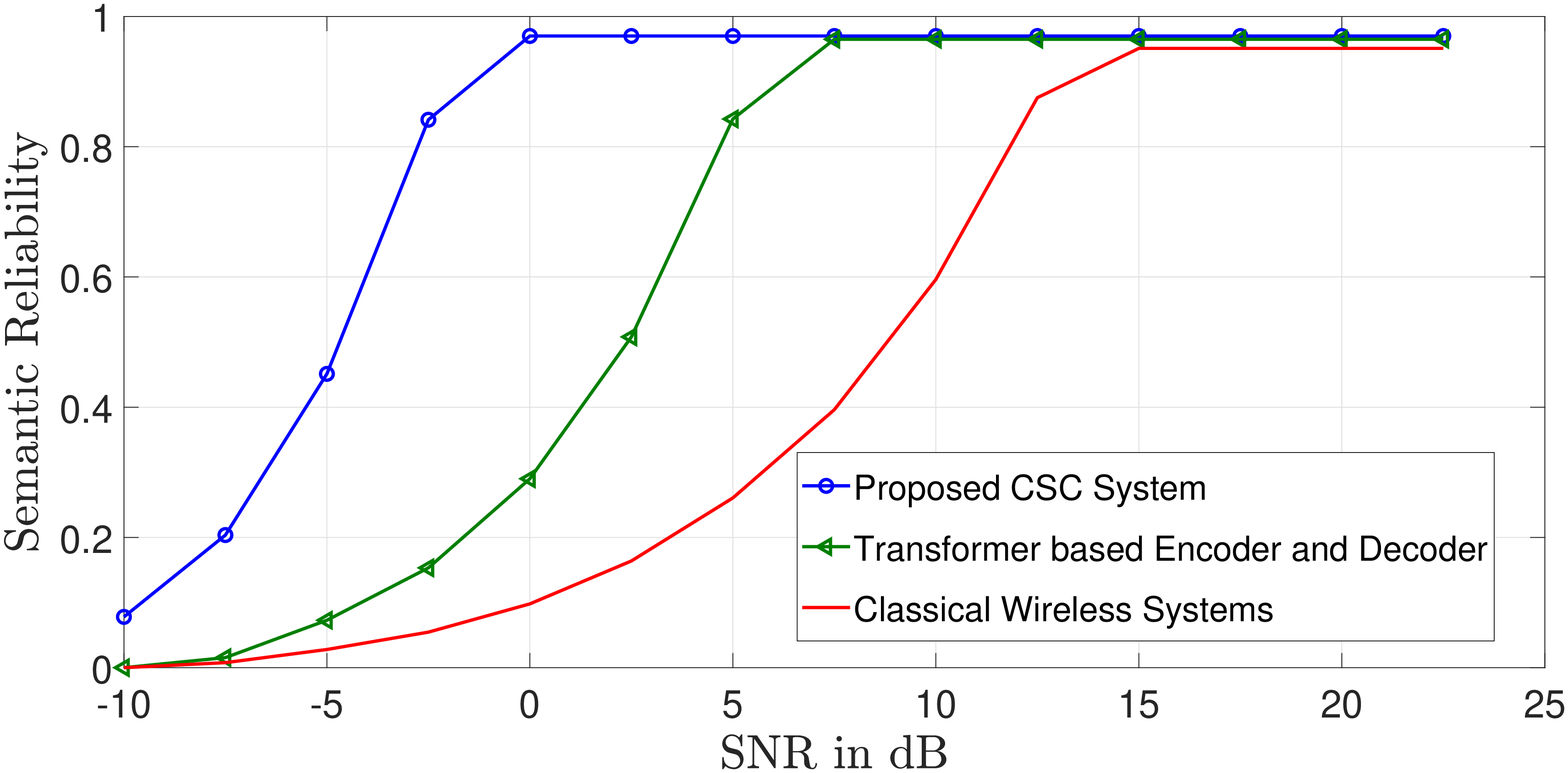}}\vspace{-3mm}
\caption{}
\label{Fig10}\vspace{-3mm}
\vspace{-2mm}
\end{subfigure}
\begin{subfigure}{0.52\textwidth}
\centering{\includegraphics[width=3.1in,height=1.5in]{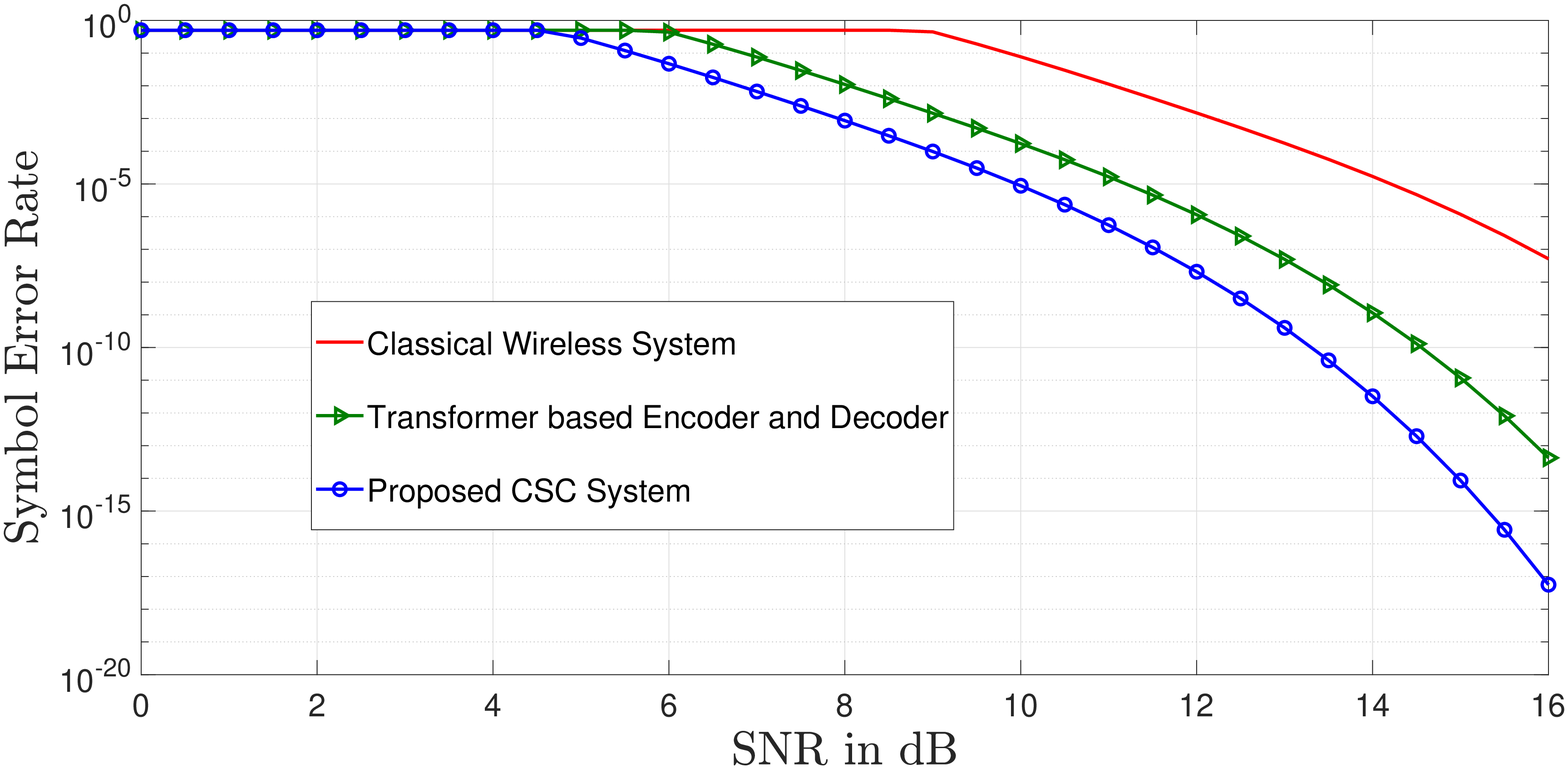}}\vspace{-3mm}
\caption{}
\label{Fig1}\vspace{-3mm}
\vspace{-2mm}
\end{subfigure}\vspace{-1mm}
\caption{\small (a) Semantic reliability vs SNR. (b) Symbol error rate vs SNR.}
\vspace{-1mm}\end{figure}
\begin{figure}[t]\vspace{-4mm}
\begin{subfigure}{0.52\textwidth}
\centering{\includegraphics[width=3.1in,height=1.5in]{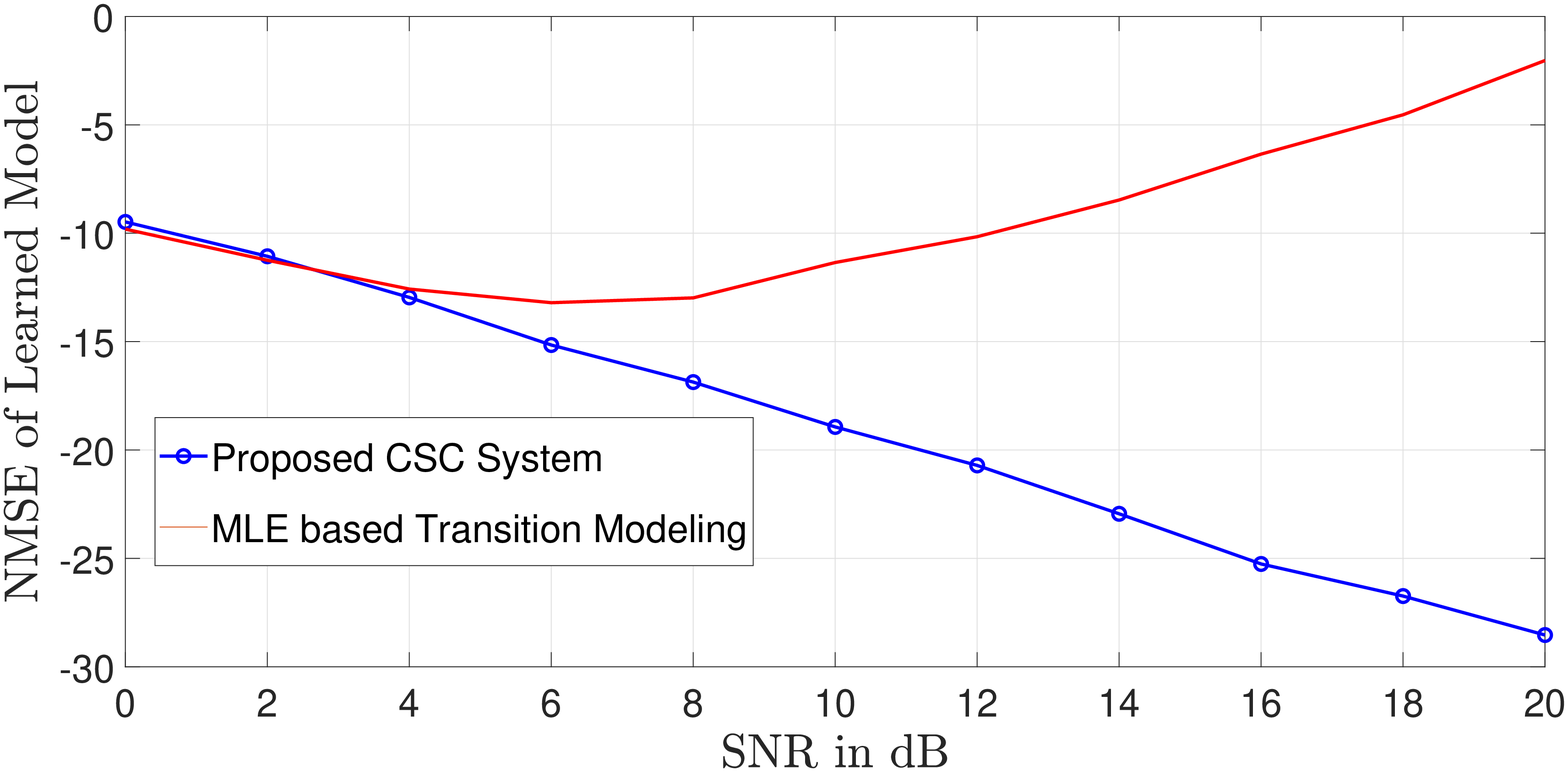}}\vspace{-3mm}
\caption{}
\label{Fig2}\vspace{-1mm}
\vspace{-2mm}
\end{subfigure}
\begin{subfigure}{0.52\textwidth}
\centering{\includegraphics[width=3.1in,height=1.5in]{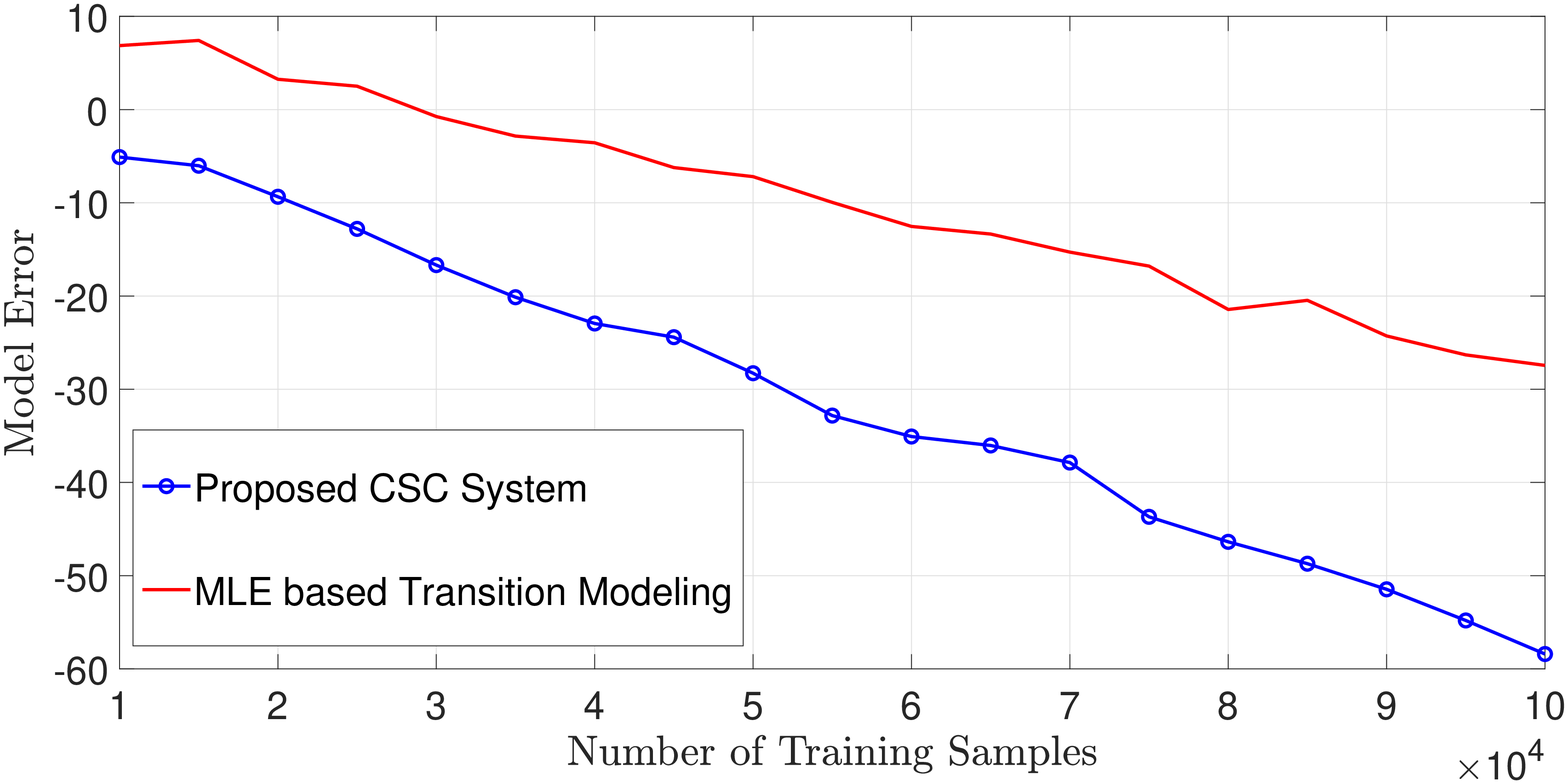}}\vspace{-3mm}
\caption{}
\label{Fig3}\vspace{-1mm}
\vspace{-2mm}
\end{subfigure}
\label{Fig_MI}\vspace{-8mm}
\caption{\small (a) Environment model error between expert and imitator agents. (b) Generalization performance.}
\end{figure}
\begin{figure}[t]
\begin{subfigure}{0.52\textwidth}
\centering{\includegraphics[width=3.1in,height=1.5in]{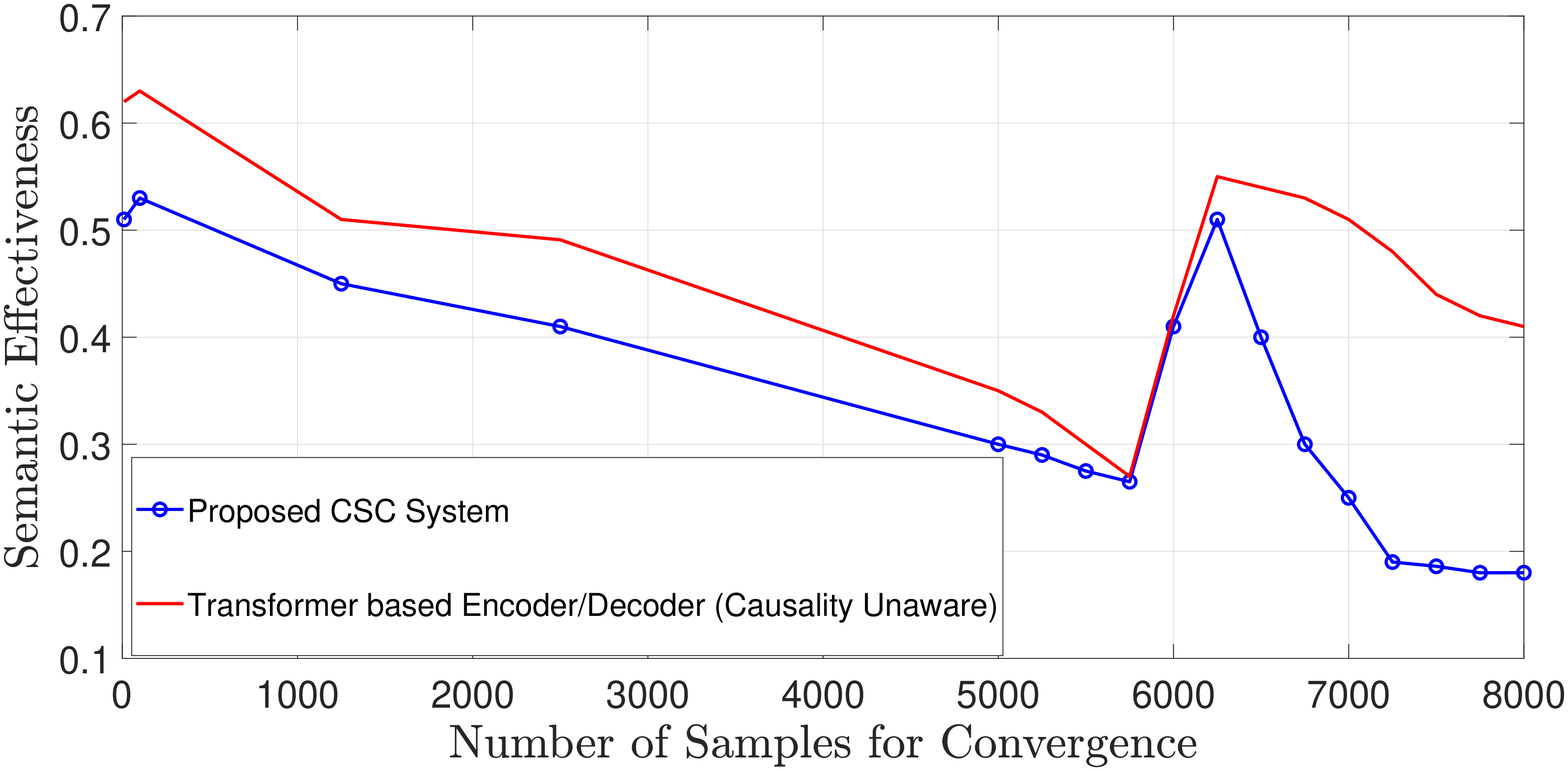}}\vspace{-3mm}
\caption{}
\label{Fig11}\vspace{-1mm}
\vspace{-2mm}
\end{subfigure}
\begin{subfigure}{0.52\textwidth}
\centering{\includegraphics[width=3.1in,height=1.5in]{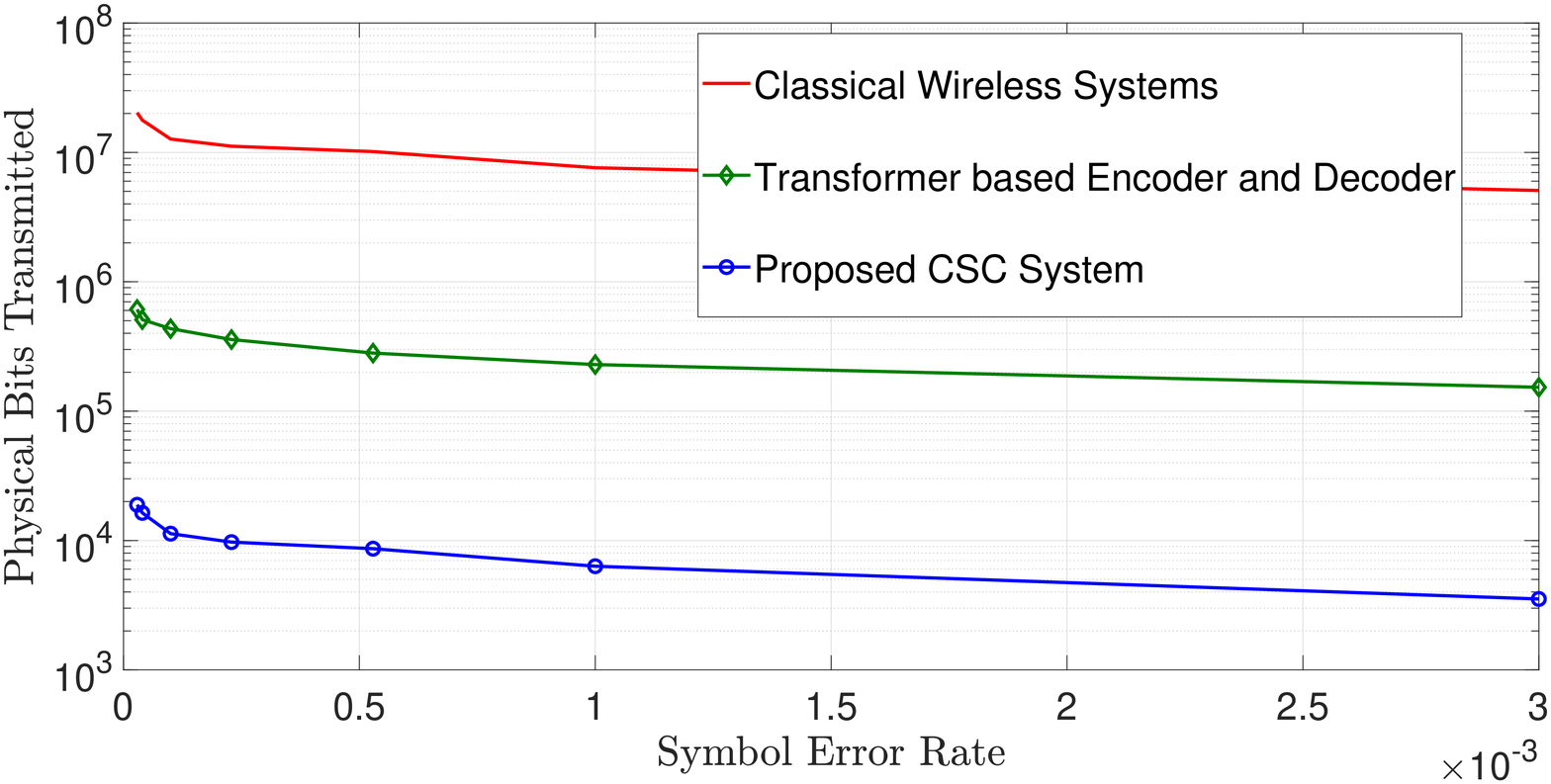}}\vspace{-3mm}
\caption{}
\label{Fig12}\vspace{-1mm}
\vspace{-2mm}
\end{subfigure}\vspace{-4mm}
\caption{\small (a) Semantic effectiveness under non-stationary learning environments (corresponding to change in dynamics $\bmg_s$).  (b) Number of physical bits to be transmitted to achieve a particular SER (corresponding to same semantic reliability).}
\vspace{1mm}\end{figure}
Fig.~\ref{FigIntegInfo} presents a scatter plot of the integrated information in the data, which was calculated using \eqref{eq_Iphi} and taking into account two-dimensional states $\bms_i^t$. The brightness of each point in the plot corresponds to the level of integrated information. Additionally, the second row of the same figure shows the SCEs identified through the algorithm described in Section~\ref{alg_SCE}. The number of SCEs comprises approximately two-thirds of the total number of states present in the data, representing a significant reduction in the amount of data that does convey any semantic information. Fig.~\ref{FigConfoundingError} illustrates that as communication progresses, the error in confounding variable estimation decreases, in contrast to a model-based Bayesian approach that experiences a performance floor of approximately $-5$ dB due to inaccurate modeling of the physical environment.

Fig.~\ref{Fig10} illustrates the superior performance of the proposed CSC system, which achieves a semantic reliability of $0.95$ at $0$ dB. In contrast, the SC system that does not consider causal structure can achieve the same level of reliability only at $7$ dB. This is attributed to the fact that the proposed CSC system requires fewer samples to achieve the desired reliability on the test data set compared to the SC system, which fails to leverage causality. Fig.~\ref{Fig1} shows the evaluation results for symbol error rate across different baselines. The results indicate that the proposed CSC system significantly outperforms the traditional SC system, achieving a $10$ dB decrease in SER (in dB) at an SNR of $8$ dB. Furthermore, compared to classical wireless systems, the proposed CSC system demonstrates a $40$ dB decrease in SER, highlighting its superior capabilities in achieving higher data rates and lower latency. 

Fig.~\ref{Fig2} presents the evaluation results for normalized mean squared error (NMSE) in the learned model between expert and imitator agents. The results demonstrate that the proposed CSC system significantly outperforms the maximum likelihood (MLE) baselines that utilizes a linear approximation \cite{TomarArxiv2021} (with just a history of length $1$) of the autoregressive model assumed for the dataset, achieving higher model accuracy across all SNRs. This is particularly important for complex datasets, where linear models may perform poorly due to potential nonlinearities. These findings highlight the importance of improving physical model accuracy using advanced AI algorithms, such as causal discovery. In Fig.~\ref{Fig3}, we evaluate the generalization performance of the proposed MBRL scheme at the imitator compared to MLE based learner. Here, we computed the model error ($\bmg_s^E - \bmg_s^{\eta}$) between expert and imitator agents for the proposed CSC and the MLE. We observe that our MBRL performs
much better than the standard model learner (MLE), especially
when the number of samples available is low. This is attributed to the causal structure extraction of the environment dynamics, which leads to better generalizability.  

Fig.~\ref{Fig11} shows the performance of the proposed CSC system and the causality unaware baseline in a non-stationary learning environment. The goal here is to demonstrate the quick adaptability (with minimal training effort compared to state-of-the-art) of the CSC approach proposed to changes in the transition dynamics $\bmg_s$ of the time series data used in the simulations. In this case, the transition dynamics $\bmg_s$ change after around $5.7$K samples. The results demonstrate that the proposed CSC system efficiently recovers from dynamics perturbations, requiring approximately three times fewer samples compared to the causality-unaware system, which necessitates more retraining efforts. Hence, the proposed approach is effective for generalizing to multiple environments. In Fig.~\ref{Fig12}, we observe that the CSC system outperforms conventional communication systems and other SC baselines in terms of efficiency (measured in the number of bits required to convey the same amount of semantic information). Specifically, we analyze the number of bits transmitted to achieve a particular SER for $10,000$ communication instances. The CSC system requires significantly fewer bits (by a factor of $1000$) compared to the classical system (without reasoning part), highlighting the significance of our proposed approach. Furthermore, when compared to state-of-the-art algorithms that do not consider causal reasoning, the proposed CSC approach is more robust to SER.

\vspace{-4mm}\section{Conclusion}
\vspace{-2mm}

In this paper, for the first time in the literature, we have presented a new vision of a DT-based SC system entitled CSC that relies upon the recently emerged theory of consciousness measures based on IIT and AI tools such as IL and MBRL. This approach enables causal discovery of the network state, allowing it to be generalized across multiple wireless environments. We
have formulated a bi-level optimization based on variational inference and information bottleneck principle to learn causal discovery, state transitions, and semantic representation at the transmitter (expert agent). The optimized semantic representation is such that the receiver is able to extract maximum semantic information while at the same time revealing minimum information about the state. At the receiver (the imitator node in IL), using a generative AI architecture originally proposed for ``world models" in RL, the node improves its knowledge about the network state transitions and causality over time. The proposed bi-level optimization is formulated using the principles of variational inference and IDS. We have shown analytically the performance shortfall associated with a suboptimal environment learning at the imitator. Simulation results demonstrate our proposed CSC's superiority in improving communication efficiency (minimal transmission) and reliability
compared to classical communication and state-of-the-art SC systems.

\vspace{-4mm}
\bibliographystyle{IEEEbib}
\def\baselinestretch{0.9} \vspace{-2mm}
\bibliography{icml2021,semantic_refs}

\newpage\section*{Supplementary Material}

\vspace{-0mm}\appendices
\vspace{-0mm}\section{Simplical Homology}
\label{Homology}

\vspace{-0mm}\begin{definition}
A $p$-simplex $\sigma$ is the convex hull of $p + 1$
affinely independent points $x_0, x_1,...,x_p \in R^d$. We denote
$σ = \textrm{conv}\{x_0,\cdots,x_p\}$. The dimension of $σ$ is $p$.
\vspace{-1mm}\end{definition}
\vspace{-1mm}\begin{definition}
A simplicial complex $\mK$ is a finite collection of simplices such that $\sigma \in \mK$ and $\tau$ being a face of $\sigma$ implies $\tau \in \mK$, and $\sigma, \sigma^{\prime} \in \mK$ implies $\sigma \cap \sigma^{\prime}$ is either empty or a face
of both $\sigma$ and $\sigma^{\prime}$.
\vspace{-1mm}\end{definition}
\vspace{-1mm}\begin{definition}
A face of $\sigma$ is $conv(S)$ where $S \subset \{x_0, \cdots , x_p\}$
is a subset of the $p + 1$ vertices.
\end{definition}
\vspace{-1mm}\begin{definition}
The total variation distance between two distributions $P$ and $\widehat{P}$, defined on domain $\mD$ can be defined as
\vspace{-1mm}\beq
D_{TV}(P,\widehat{P}) = \sup\limits_{x\in \mD} \abs{P(x) - \widehat{P}(x)}.
\vspace{-1mm}\eeq
\vspace{-1mm}\end{definition}

\vspace{-1mm}\section{Proof of Lemma~\ref{lemma_intrinsicInfo}}
\label{lemma_intrinsicInfo_proof}
\vspace{-1mm}

We can write $p\left(\bms^{t-1}\mid \bms^t\right) = \sum\limits_{\btheta} p\left(\btheta\right)p\left(\bms^{t+1}\mid \bms^t,\btheta\right)$. Therefore, for the imitator
\vspace{-2mm}\beq \small
\mathbb{I}_c^{\eta} = \sum\limits_{\btheta} p\left(\btheta\mid \bmH_t\right)\mathbb{D}\left(p\left(\bms^{t-1}\mid \bms^t,\btheta\right ) \mid\mid p\left(\bms^{t-1}\right)\right) \leq \mathbb{D}\left(p\left(\bms^{t-1}\mid \bms^t,\btheta^0\right ) \mid\mid p\left(\bms^{t-1}\right)\right),\vspace{-2mm} \label{eq_Ic_eta}\eeq where $\btheta^0$ represents the accurate $\btheta^0$ that is observed by the expert agent. The proof for effect information follows similar lines. Hence, $\mathbb{I}_c^{\eta} \leq \mathbb{I}_c^{E}$ and $\mathbb{I}_e^{\eta} \leq \mathbb{I}_e^{E}$, which means that $\mathbb{I}_{ce}^{\eta} \leq \mathbb{I}_{ce}^{E}$. Further, we can rewrite $
\mathbb{I}_{ce}^{\eta} - \mathbb{I}_{ce}^{E} $ as follows.
\beq
\begin{aligned}
\mathbb{I}_{ce}^{\eta} - \mathbb{I}_{ce}^{E}  &= \abs{\sum\limits_{\btheta} p\left(\btheta\mid \bmH_t\right)\mathbb{D}\left(p\left(\bms^{t-1}\mid \bms^t,\btheta\right ) \mid\mid p\left(\bms^{t-1}\right)\right)- \sum\limits_{\btheta}\delta_{\btheta^0}(\btheta)\mathbb{D}\left(p\left(\bms^{t-1}\mid \bms^t,\btheta^0\right ) \mid\mid p\left(\bms^{t-1}\right)\right)} \\ 
&=\left\lvert\sum\limits_{\btheta} p\left(\btheta\mid \bmH_t\right)\mathbb{D}\left(p\left(\bms^{t-1}\mid \bms^t,\btheta\right ) || p\left(\bms^{t-1}\right)\right)- \delta_{\btheta^0}(\btheta)\mathbb{D}\left(p\left(\bms^{t-1}\mid \bms^t,\btheta^0\right ) || p\left(\bms^{t-1}\right)\right)\right\rvert \\
&\stackrel{(a)} \leq \abs{\sum\limits_{\btheta} \mathbb{D}\left(p\left(\bms^{t-1}\mid \bms^t,\btheta\right ) || p\left(\bms^{t-1}\right)\right)}\epsilon = \mathbb{I}_c(\bms^{t-1})\epsilon = \mO(\epsilon),
\end{aligned}\label{eq_err_intrinsic_info}
\eeq
where $(a)$ follows from the assumption that $D_{TV}(p(\btheta\mid\bmH_t)\mid\mid\delta_{\btheta^0}(\btheta)) \leq \epsilon$. \eqref{eq_err_intrinsic_info} shows that as long as the intrinsic information provided by any state $\bms^{t}$ is finite, the error intrinsic information between expert and imitator agents can stay proportional to or below $\mO(\epsilon)$.

\vspace{-1mm}\section{Proof of Lemma~\ref{lemma_err_semInfo}}
\label{app_lemma_err_semInfo}
\vspace{-2mm}

The error in integrated information can be written as 
\vspace{-3mm}\beq\small
\mathbb{I}_{\phi,c}^{\eta} - \mathbb{I}_{\phi,c}^{E} = I_c(\bms^{t-1};\bms^t,\btheta) - I_c(\bms^{t-1};\bms^t,\widehat{\btheta}).
\label{eq_err_mi}
\vspace{-3mm}\eeq
We can rewrite \eqref{eq_err_mi} as
$\small
\mathbb{I}_{\phi,c}^{\eta} - \mathbb{I}_{\phi,c}^{E} = H\left(\bms^{t-1}\mid \bms^t,\bthetah(\bmH_t)\right) - H\left(\bms^{t-1}\mid \bms^t,\btheta\right).
$
Further, we look at an averaged error measure (averaged across several estimates of $\bthetah$), which can be written as:
\vspace{-3mm}\beq\small
\begin{array}{l}
\mathbb{I}_{\phi,c}^{\eta} - \mathbb{I}_{\phi,c}^{E} = \sum\limits p\left(\btheta = \bthetah\mid \bmH_t\right)p(\bms^{t-1}\mid \bms^{t},\bthetah)\log p\left(\bms^{t-1}\mid \bms^t,\bthetah(\bmH_t)\right)  - H\left(\bms^{t-1}\mid \bms^t,\btheta\right). 
\end{array}
\vspace{-3mm}\eeq
To simplify this further, we can write $\frac{p(\bms^{t-1},\bms^t,\bthetah)}{p(\bms^t,\bthetah)} = p(\bms^{t}\mid \bms^{t-1},\bthetah)$.
\vspace{-3mm}\beq\small
\begin{aligned}
\mathbb{I}_{\phi,c} - \mathbb{I}_{\phi,c}^e & = \sum\limits p\left(\btheta = \bthetah\mid \bmH_t\right) p(\bms^{t-1}\mid \bms^{t},\bthetah) \log p\left(\bms^{t-1}\mid \bms^t,\bthetah(\bmH_t)\right)  - H\left(\bms^{t-1}\mid \bms^t,\btheta\right) \\ 
& = \sum\limits p\left(\btheta = \bthetah\mid \bmH_t\right) p(\bms^{t}\mid \bms^{t-1},\bthetah) \log p\left(\bms^{t}\mid \bms^{t-1},\bthetah(\bmH_t)\right)  - H\left(\bms^{t-1}\mid \bms^t,\btheta\right) \\
& = \mathbb{E}_{q}\left(H\left(\bms^{t-1}\mid \bms^t,\btheta\right) - H\left(\bms^{t-1}\mid \bms^t,\btheta)\right)\right).
\end{aligned}
\vspace{-3mm}\eeq

\vspace{-4mm}\section{Proof of Theorem~\ref{def_topos}}
\label{def_topos_proof}
\vspace{-3mm}

We prove the transitive property by a counterexample. Let us suppose that $\bmd_1\prec_{w_1}\bmd_2$ and $\bmd_2\prec_{w_2}\bmd_3$, where $\phi_{r_1}>0$ and $\phi_{r_2}>0$. Also, assume that $\bmd_1$ is not related to $\bmd_3$. This implies that there is an overlapping purview between $\bmd_2$ and $\bmd_3$ that is not related to $\bmd_1$. Let this partition be denoted by $\bmd_x$, which is a subset of $\bmd_2$.
We can then partition $\bmd_2$ into two sets: $\bmd_x$ and $\bmd_y$, where $\bmd_x\cap \bmd_y = \emptyset$. Note that $\bmd_x$ is the portion of $\bmd_2$ that does not have any overlap with $\bmd_1$. Since the irreducibility measure of the overlapping purview between $\bmd_2$ and $\bmd_3$ is non-zero, it follows that the irreducibility measure of $\bmd_x$ is also non-zero. Now, we can conclude that $\bmd_2$ is reducible since it can be partitioned into two subsets with non-zero irreducibility measures. Furthermore, the integrated information of a reducible system is zero. Hence, we can conclude that $\bmd_2$ does not exist as a causal distinction or a concept. Therefore, we have proven that if $\bmd_1\prec_{w_1}\bmd_2$ and $\bmd_2\prec_{w_2}\bmd_3$, then $\bmd_1$ is causally related to $\bmd_3$. Otherwise, if $\bmd_1$ is not causally related to $\bmd_3$, then $\bmd_2$ does not exist as a causal distinction. The monotonicity property follows directly from the assumption that the number of causal distinctions in $\bmd_1$ is less than that in $\bmd_2$.

\vspace{-5mm}\section{Proof of Lemma~\ref{theorem1}}
\label{theorem1_proof}
\vspace{-3mm}

We use $V^{\pi}(\bms,\mE)$ and $V^{\pi}(\bms,\mEh)$ to refer to the value of policy $\pi$ when starting from any state $\bms$ in the MDP $\mE$ and its variant $\mEh$, respectively. To simplify the notation, we introduce the following definition.
\vspace{-3mm}\beq
P^{\pi_{\eta}}_{\mE}(\cdot\mid\bms) := \mathbb{E}_{\bma \mid \pi(\cdot\mid\bms)}P^{\pi_{\eta}}_{\mE}(\cdot\mid\bms,\bma)\,\,\, \mbox{and}\,\,\, P^{\pi_{\eta}}_{\mEh}(\cdot\mid\bms) := \mathbb{E}_{\bma \mid \pi(\cdot\mid\bms)}P^{\pi_{\eta}}_{\mEh}(\cdot\mid\bms,\bma)
\vspace{-3mm}\eeq
Before the proof, we note the following useful observations. 
\vspace{-3mm}\begin{itemize}
\item Since $D_{TV} (P_\mE(·\mid\bms, \bma), P_{\mEh}(·\mid \bms, \bma)) \leq \epsilon_{\mE} \forall(\bms, \bma),$ the inequality also holds for an average over actions, i.e.
$D_{TV} (P_\mE(·\mid\bms), P_{\mEh}(·\mid \bms)) \leq \epsilon_{\mE} \forall \bms$.
\item Given that the rewards are bounded, it is possible to attain a maximum reward of $R_{\textrm{max}}$ at each time step. By using a geometric series with a discount factor of $\gamma$, we can express this as follows:
\vspace{-1mm}\beq
\max\limits_{\bms\in\mS} V^{\pi}(\bms,\mE) \leq \frac{R_{\textrm{max}}}{1-\gamma}\,\,\, \forall \pi, \bms
\vspace{-3mm}\eeq
\item Consider a real-valued function $f(x)$ defined over $\mX$ such that its output range is limited to $[-f_{\textrm{max}}, f_{\textrm{max}}]$.  Let $P_1(x)$ and
$P_2(x)$ be two probability distribution (density) over the space $\mX$. Then, we have
\vspace{-3mm}\beq
\abs{\mathbb{E}_{x\mid P_1(x)} f(x) - \mathbb{E}_{x\mid P_2(x)} f(x)} \leq 2f_{\textrm{max}}D_{TV}(P_1,P_2).
\vspace{-3mm}\eeq
\end{itemize}
Based on the observations mentioned above, we can establish the following set of inequalities:
\vspace{-3mm}\beq
\begin{aligned}
\abs{V^{\pi}(\bms,\mE) - V^{\pi}(\bms,\mEh)} & = \abs{\mR(\bms) + \gamma\mbE_{\bms^{\prime} \sim P^{\pi}_{\mE}(\cdot\mid \bms)} V^{\pi}(\bms^{\prime},\mE) - \mR(\bms) - \gamma\mbE_{\bms^{\prime} \sim P^{\pi}_{\mEh}(\cdot\mid \bms)} V^{\pi}(\bms^{\prime},\mEh)} \\
& \leq \gamma\abs{\mbE_{\bms^{\prime} \sim P^{\pi}_{\mE}(\cdot\mid \bms)} V^{\pi}(\bms^{\prime},\mE) - \mbE_{\bms^{\prime} \sim P^{\pi}_{\mEh}(\cdot\mid \bms)} V^{\pi}(\bms^{\prime},\mE)} \\& + \gamma\abs{\mbE_{\bms^{\prime} \sim P^{\pi}_{\mEh}(\cdot\mid \bms)} \left[V^{\pi}(\bms^{\prime},\mE) -  V^{\pi}(\bms^{\prime},\mEh)\right]} \\
& \leq2\gamma \left(\max_{\bms^{\prime}\in\mS}V^{\pi}(\bms^{\prime},\mE)\right)D_{TV}(P^{\pi}_{\mE}(\cdot\mid \bms),P^{\pi}_{\mEh}(\cdot\mid \bms)) \\ & + \gamma \left(\max_{\bms^{\prime}\in\mS}\abs{V^{\pi}(\bms^{\prime},\mE)-V^{\pi}(\bms^{\prime},\mEh)}\right)
\end{aligned}
\vspace{-1mm}\eeq
Since the above bound is applicable to all states, we have $\forall \pi$
\vspace{-1mm}\beq
\begin{aligned}
(1-\gamma)\max\limits_{\bms^{\prime}\in \mS}\abs{V^{\pi}(\bms^{\prime},\mE)-V^{\pi}(\bms^{\prime},\mEh)} & \leq \frac{2\gamma R_{\textrm{max}}}{1-\gamma}D_{TV}(P^{\pi}_{\mE}(\cdot\mid \bms),P^{\pi}_{\mEh}(\cdot\mid \bms)) \\ & \leq \frac{2\gamma \epsilon_{\mE}R_{\textrm{max}}}{1-\gamma}.
\end{aligned}
\vspace{-3mm}\eeq

\vspace{-1mm}\section{Proof of Theorem~\ref{err_gap}}
\label{err_gap_proof}
\vspace{-3mm}

The first step taken is to simplify the performance difference, followed by bounding the different terms. The proof follows similar steps as in \cite{RajeswaranPMLR2020}. Let $\pi_{\mEh}^{\prime}$ to be an optimal
policy in the model, so that $J(\pi_{\mEh}^{\prime},\mEh) \geq J(\pi_{\mEh},\mEh),\forall \pi$. We can break down the performance difference into different contributions as follows:
\vspace{-3mm}\beq
\begin{aligned}
J(\pi^{*},\mE) - J(\pi,\mE) = J(\pi^{*},\mE) - J(\pi^{*},\mEh) + J(\pi^{*},\mEh) - J(\pi,\mE) \\
= \underbrace{J(\pi^{*},\mE) - J(\pi^{*},\mEh)}_{\textrm{Term-I}} +  \underbrace{J(\pi^{*},\mEh) - J(\pi,\mEh)}_{\textrm{Term-II}} +  \underbrace{J(\pi,\mEh) - J(\pi,\mE)}_{\textrm{Term-III}}
\end{aligned}
\vspace{-2mm}\eeq
Let's focus our attention on Term-II, which pertains to sub-optimality in the planning problem. Note that $J(\pi^{*},\mEh) - J(\pi^{*},\mEh) = J(\pi^{*},\mE) - J(\pi^{*}_{\mEh},\mEh) + J(\pi^{*}_{\mEh},\mEh) - J(\pi^{*},\mEh)$. 
We have  $J(\pi^{*},\mEh) - J(\pi^{*}_{\mEh},\mEh) \leq 0$ since $\pi^{*}$ is the optimal policy in the model, and we have $J(\pi^{*}_{\mEh},\mEh) - J(\pi^{*},\mEh) \leq \epsilon$ due to the approximate equilibrium condition. For Term-III, we will refer to the model error performance difference theorem (Theorem~\ref{theorem1}). It's worth noting that the equilibrium condition of low error, along with Pinsker's inequality \cite{CsiszarCUP2011}, implies that $\mathbb{E}_{\bms \sim \mu{\mE}^{\pi}}\left[D_{TV}(P_{\mE}(\cdot\mid\bms,\bma),P_{\mEh}(\cdot\mid\bms,\bma))\right] \leq \sqrt{\epsilon_{\mE}}$. 
Using this and Theorem~\ref{theorem1}, we have
\vspace{-1mm}\beq
J(\pi,\mEh) - J(\pi,\mE) \leq \frac{2\gamma \sqrt{\epsilon_{\mE}}R_{\textrm{max}}}{(1-\gamma)^2\mathbb{I}_0t}.
\vspace{-1mm}\eeq
Lastly, Term-I is a transfer learning term that quantifies the error of $\mEh$ (which has low error under $\pi$) when operating under the distribution of $\pi^{*}$. The performance difference can be expressed as:
\vspace{-3mm}\beq
\begin{aligned}
J(\pi^{*},\mE) - J(\pi^{*},\mEh) & = \frac{1}{1-\gamma}\mathbb{E}_{(\bms,\bma)\sim\mu_{\mE}^{\pi^{*}}}\left[\mR(\bms)\right]-\frac{1}{1-\gamma}\mathbb{E}_{(\bms,\bma)\sim\mu_{\mEh}^{\pi^{*}}}\left[\mR(\bms)\right] \\& \leq \frac{2R_{\textrm{max}}}{(1-\gamma)\mathbb{I}_0t}D_{TV}(\mu_{\mE}^{\pi},\mu_{\mEh}^{\pi^{*}}).
\end{aligned}
\vspace{-3mm}\eeq
Putting all the terms together, we have
\vspace{-3mm}\beq
J(\pi^{*},\mE) - J(\pi,\mE)  \leq \frac{2R_{\textrm{max}}}{(1-\gamma)\mathbb{I}_0t}D_{TV}(\mu_{\mE}^{\pi^{*}},\mu_{\mEh}^{\pi^{*}}) + \epsilon_{\pi} + \frac{2\gamma \sqrt{\epsilon_{\mE}}R_{\textrm{max}}}{(1-\gamma)^2\mathbb{I}_0t}.
\vspace{-3mm}\eeq

\vspace{-3mm}\section{Proposed Solution for the Bi-level Optimization}
\label{BilevelOpt}
\vspace{-3mm}

The proposed bi-level optimization solution follows a similar procedure to that of Stackelberg games \cite{BassarSpringer2018}. To begin, we briefly describe the strategy for continuous bi-level optimization. Consider a two-level optimization problem involving problems $A$ and $B$, with their respective parameters denoted by $\btheta_A$ and $\btheta_B$. The objective function for problem $A$ is represented by $L_A(\btheta_A, \btheta_B)$, and that for problem $B$ is represented by $L_B(\btheta_A, \btheta_B)$, and both aim to minimize their respective losses. The bi-level optimization problem can be solved usings a nested optimization procedure:
\vspace{-3mm}\begin{equation}
\begin{aligned}
&\min_{\btheta_A} L_A(\btheta_A, \btheta_B^*(\btheta_A)) \
&\text{subject to} \quad \btheta_B^*(\btheta_A) = \min_{\btheta_B} L_B(\btheta_A, \btheta_B),
\end{aligned}
\vspace{-2mm}\end{equation}
where the expression $\btheta_B^*(\btheta_A)$ denotes the best solution for problem $B$ given the outer level problem $A$'s choice of $\btheta_A$, and $\btheta_A$ represents the solution to the optimization problem described earlier. In the inner level, problem $B$ implicitly selects its parameters based on the choice of $\btheta_A$, and problem $A$ is aware of this relationship and can use this information to update its own parameters.

To solve the nested optimization, we can focus on optimizing $\btheta_A$ iteratively. Specifically, we can update $\btheta_A$ as: $\btheta_A \leftarrow \btheta_A \!-\! \alpha_A \frac{\partial L_A(\btheta_A, \btheta_B(\btheta_A^*))}{\partial \btheta_A}$, where $\alpha_A$ is the learning rate and the gradient can be written as:
\vspace{-2mm}\beq
\begin{array}{l}
\frac{\partial L_A(\btheta_A, \btheta_B(\btheta_A^*))}{\partial \btheta_A} = \frac{\partial \btheta_B^*(\btheta_A)}{\partial \btheta_A} \frac{\partial L_A(\btheta_A, \btheta_B(\btheta_A^*))}{\partial \btheta_B} \mid_{ \btheta_B = \btheta_B^*} + \frac{\partial L_A(\btheta_A, \btheta_B(\btheta_A^*))}{\partial \btheta_A} \mid_{ \btheta_B = \btheta_B^*}
\end{array}
\vspace{-2mm}\eeq
The Jacobian $\frac{\partial \btheta_B^{*}}{ (\partial\btheta_A)}$ can be computed using the implicit function theorem as in \cite{KrantzSpringer2003}. Thus, in principle, we can compute the gradient with respect to the $A$'s parameters and solve the nested optimization, to at least a local minimizer.


\end{document}